\documentclass[runningheads]{llncs}

\usepackage{eccv}

\usepackage{eccvabbrv}

\usepackage{graphicx}
\usepackage{booktabs}

\usepackage[pagebackref,breaklinks,colorlinks,citecolor=eccvblue]{hyperref}

\usepackage{color,xcolor}
\usepackage{epsfig}
\usepackage{graphicx}

\usepackage{adjustbox}
\usepackage{array}
\usepackage{booktabs}
\usepackage{colortbl}
\usepackage{float,wrapfig}
\usepackage{hhline}
\usepackage{multirow}
\usepackage{subcaption}
\usepackage[font={small}]{caption}

\usepackage{bm}
\usepackage{nicefrac}
\usepackage{microtype}
\usepackage{mathtools}

\usepackage{changepage}
\usepackage{extramarks}
\usepackage{fancyhdr}
\usepackage{lastpage}
\usepackage{setspace}
\usepackage{soul}
\usepackage{xspace}

\usepackage{url}

\usepackage{algorithmic}
\usepackage{algorithm}
\usepackage{enumitem}
\usepackage{bbm}

\usepackage{bm}

\def\eqref#1{equation~\ref{#1}}

\def\1{\bm{1}}

\DeclareMathAlphabet{\mathsfit}{\encodingdefault}{\sfdefault}{m}{sl}
\SetMathAlphabet{\mathsfit}{bold}{\encodingdefault}{\sfdefault}{bx}{n}

\newcommand{\model}{MagicMirror\xspace}

\newcommand{\menglei}[1]{{\color{blue}[Menglei: #1]}}
\newcommand{\PG}[1]{{\color{blue}[PG: #1]}}
\newcommand{\sergio}[1]{{\color{blue}[sergio: #1]}}
\definecolor{gold}{rgb}{0.7, 0.5, 0}

\newcommand{\MM}[1]{{\color{gold}{\bf [MM: #1]}}}

\newcommand{\MB}[1]{{\color{green}[Marcel: #1]}}

\renewcommand{\menglei}[1]{}
\renewcommand{\PG}[1]{}
\renewcommand{\sergio}[1]{}

\renewcommand{\MM}[1]{}
\renewcommand{\MB}[1]{}

\begin{document}

\title{MagicMirror: Fast and High-Quality Avatar Generation with a Constrained Search Space} 

\titlerunning{MagicMirror}

\author{Armand Comas-Massagu\'e\inst{1,2} \and
Di Qiu\inst{1} \and
Menglei Chai\inst{1} \and
Marcel B\"uhler\inst{1,3} \and \\
Amit Raj\inst{1} \and
Ruiqi Gao\inst{4} \and
Qiangeng Xu\inst{1} \and
Mark Matthews\inst{1} \and
Paulo Gotardo\inst{1} \and
Octavia Camps\inst{2} \and
Sergio Orts-Escolano\inst{1} \and
Thabo Beeler\inst{1}}

\authorrunning{A. Comas et al.}
\institute{Google \and Northeastern University \and ETH Z\"urich \and Google DeepMind}

\maketitle
\vspace{-0.4cm}
\begin{figure*}[ht]
\begin{center}
\includegraphics[width=\textwidth]{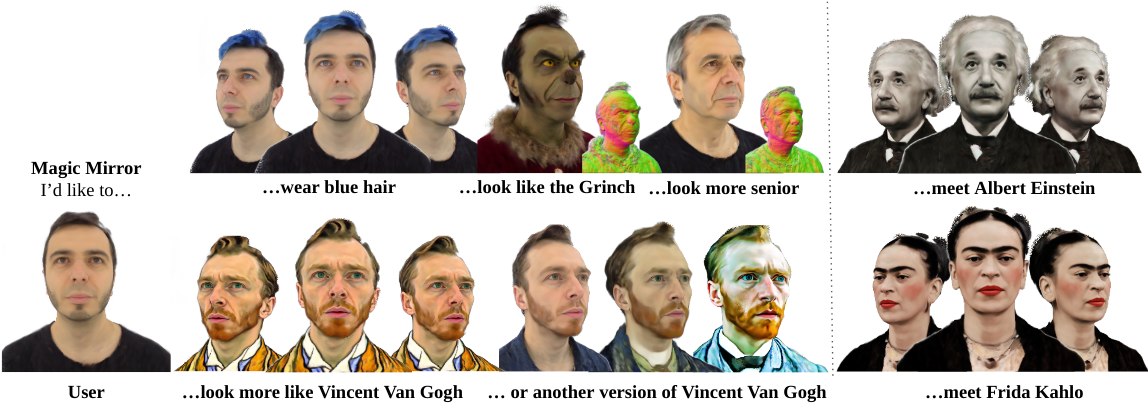}
\vspace{-13pt}
\caption{\label{fig:teaser} We propose MagicMirror, a method for fast text-guided 3D avatar head generation, with the option of subject personalization. (left) We illustrate how given subject pictures, MagicMirror can generate a 3D avatar with the subject's stylized appearance by following text descriptions. Avatars exhibit high-quality in geometry and texture, with significant altered while preserving the identity of the subject. (right) It can also generate well-known characters by only employing a text prompt. }
\end{center}
\vspace{-1.1cm}
\end{figure*}

\begin{abstract}
  We introduce a novel framework for 3D human avatar generation and personalization, leveraging text prompts to enhance user engagement and customization. Central to our approach are key innovations aimed at overcoming the challenges in photo-realistic avatar synthesis. Firstly, we utilize a conditional Neural Radiance Fields (NeRF) model, trained on a large-scale unannotated multi-view dataset, to create a versatile initial solution space that accelerates and diversifies avatar generation. Secondly, we develop a geometric prior, leveraging the capabilities of Text-to-Image Diffusion Models, to ensure superior view invariance and enable direct optimization of avatar geometry. These foundational ideas are complemented by our optimization pipeline built on Variational Score Distillation (VSD), which mitigates texture loss and over-saturation issues. As supported by our extensive experiments, these strategies collectively enable the creation of custom avatars with unparalleled visual quality and better adherence to input text prompts. You can find more results and videos in our website: \href{https://syntec-research.github.io/MagicMirror}{syntec-research.github.io/MagicMirror}
\end{abstract}

\section{Introduction}\label{sec:intro}
\begin{figure}[t!]
\centering
  \includegraphics[ trim={0 130 70 0}, width=1\textwidth]{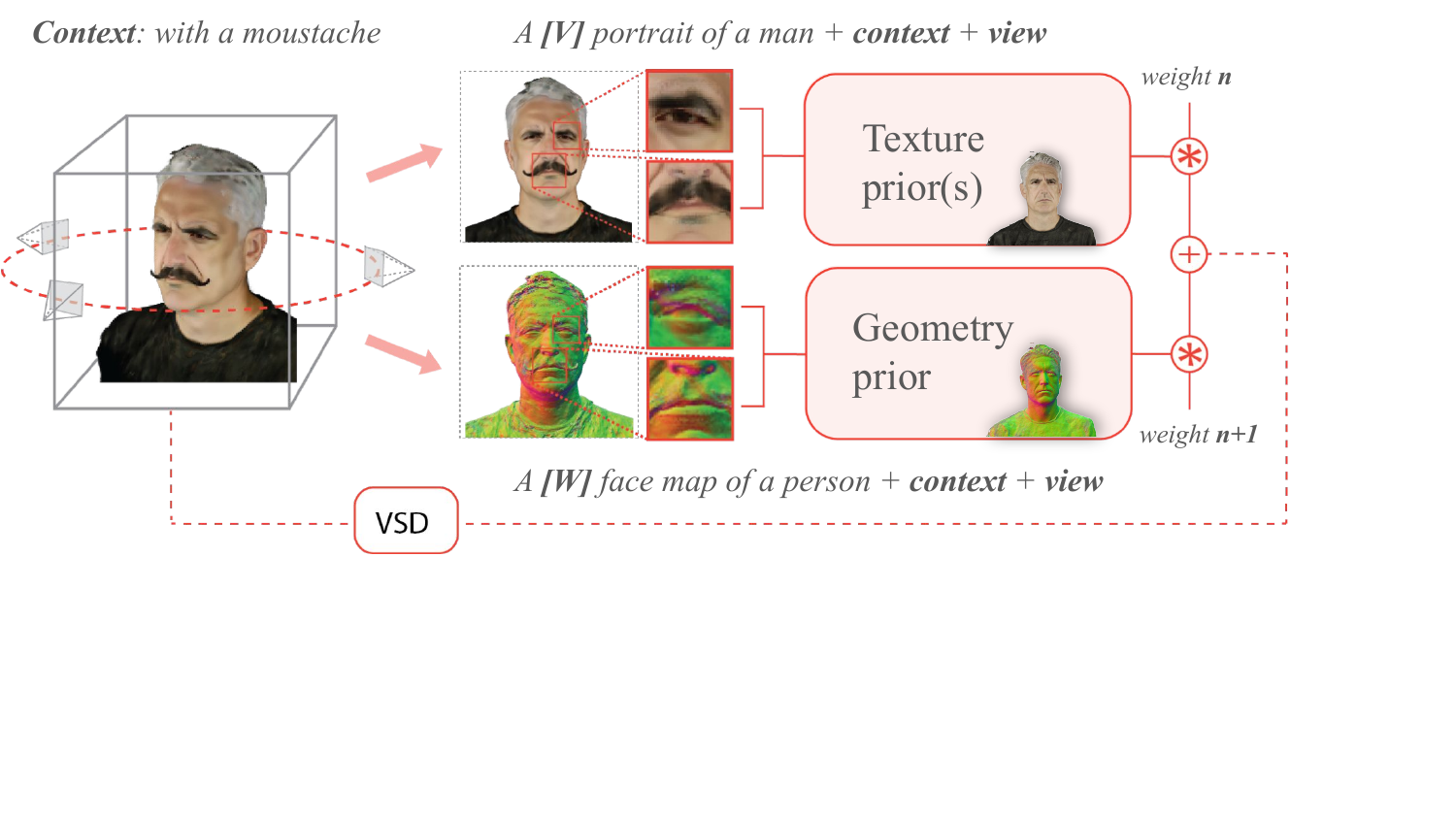}
    \caption{Our two pipelines for 3D head avatar generation and customization follow the same structure: a pre-trained conditional NeRF model serves as 3D prior for fast avatar generation. Our pipelines additionally leverage two pre-trained text-to-image diffusion models as texture and geometry priors, allowing for distillation-based customization of both these components based on input text prompts with state-of-the-art quality.}
\label{fig:arch}
\end{figure}

Customizable 3D human avatars are central for many experiences such as gaming, v-tubing, augmented and virtual reality (AR/VR), or telepresence applications.
Intuitive editing and personalization techniques for such avatars are highly desirable, as customized avatars provide a greater sense of engagement, ownership and aid adoption of the aforementioned technologies. 
Traditional CGI editing techniques, however, are still difficult, non-intuitive, and laborious for the average user. Recently, text-prompting has emerged as a natural and intuitive interface to control the creation and customization of highly complex generative outputs, due to the impressive progress of Language-Image modeling \cite{radford2021learning} and Text-to-Image Diffusion Models \cite{Ho2020ddpm}.
Two main approaches have emerged for generative modeling of 3D assets: direct 3D modeling, and neural rendering techniques leveraging 2D images.

Direct 3D models largely conform to the text-to-image paradigm which train a generative model from a large dataset of labeled 3D assets\cite{huang2023humannorm}. 

Sourcing such data at scale, however, is difficult and expensive\cite{Deitke023objaverse}. 
3D assets are nowhere nearly as abundant as the 2D images easily available on the Internet. 
Furthermore, 3D assets that are available typically lack the rich semantic information that often accompanies Internet images. 
Consequently, the results from this category typically lack diversity and quality compared to their 2D large-scale counterparts.

The second category of methods leverage the implicit 3D knowledge within 2D generative models, lifting 2D outputs onto 3D via differential rendering and novel objective functions.
Several designs of these objective functions have been proposed, including simple reconstruction losses based on transformed 2D images \cite{Haque2023inerf2nerf}, high-level text-image misalignment scores \cite{Wang2023nerfart}, and model distillation \cite{poole2023dreamfusion}.

These methods work best if the outputs of the 2D model are multi-view consistent, which is usually not the case, leading to non-convergence and infamous "Janus face" artifacts~\cite{poole2023dreamfusion}.
Although appealing, this second category of methods has several key challenges.

First, despite their ability to generate large amounts of text-guided 2D images and supervisory signals, they are not guaranteed to be multi-view consistent.

Consequently, 3D optimization suffers from conflicting supervision.

This issue can be mitigated by reducing the amount of overlap between views, through reduced view count, and evenly distributing views. However, a possible unfortunate effect is in making the problem ill-posed, thus leading to poor results.

An approach to improve reconstructions might then be to improve the multi-view consistency of existing 2D image generators.

This could be achieved by training on multi-view data and sharing information across views \cite{shi2023mvdream,liu2023syncdreamer}.

While hopeful, this approach still bears the burden of sourcing multi-view data, often as difficult as sourcing 3D assets described above.

Other approaches based on model distillation require the use of a high classifier-free guidance weight \cite{ho2022classifier}, causing textureless and over-saturated results \cite{poole2023dreamfusion} and, more importantly, reducing diversity \cite{wang2023prolificdreamer}.

\begin{figure*}[t!]
\begin{center}
\small
\newcommand{\height}{1.5cm}
\begin{tabular}{cccccc}
    \includegraphics[height=\height]{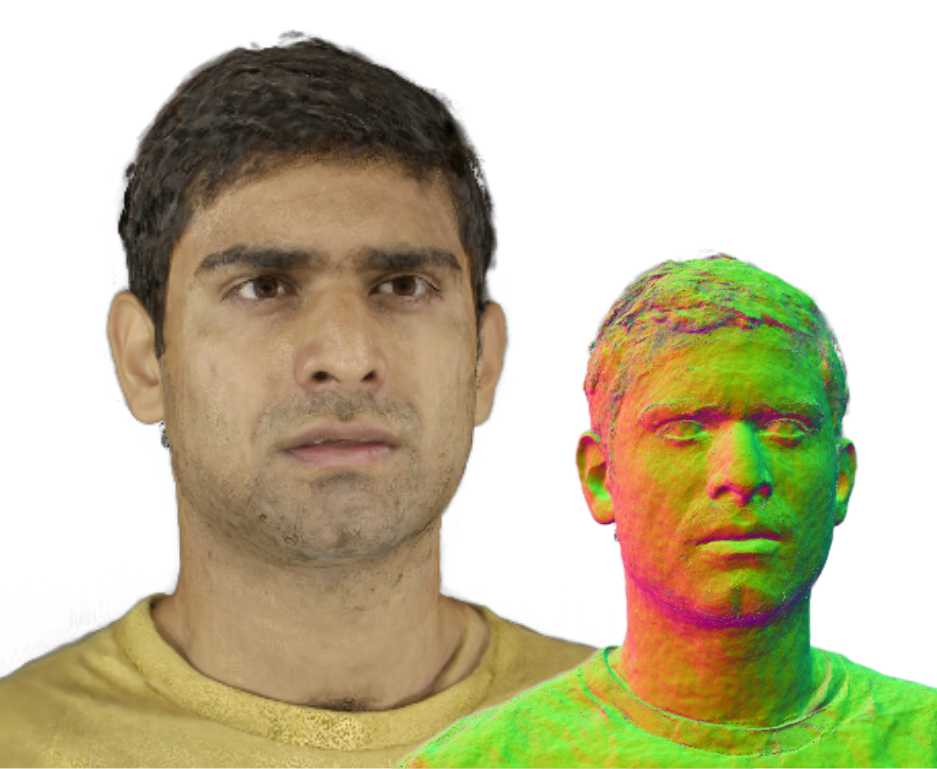} &
    \includegraphics[height=\height]{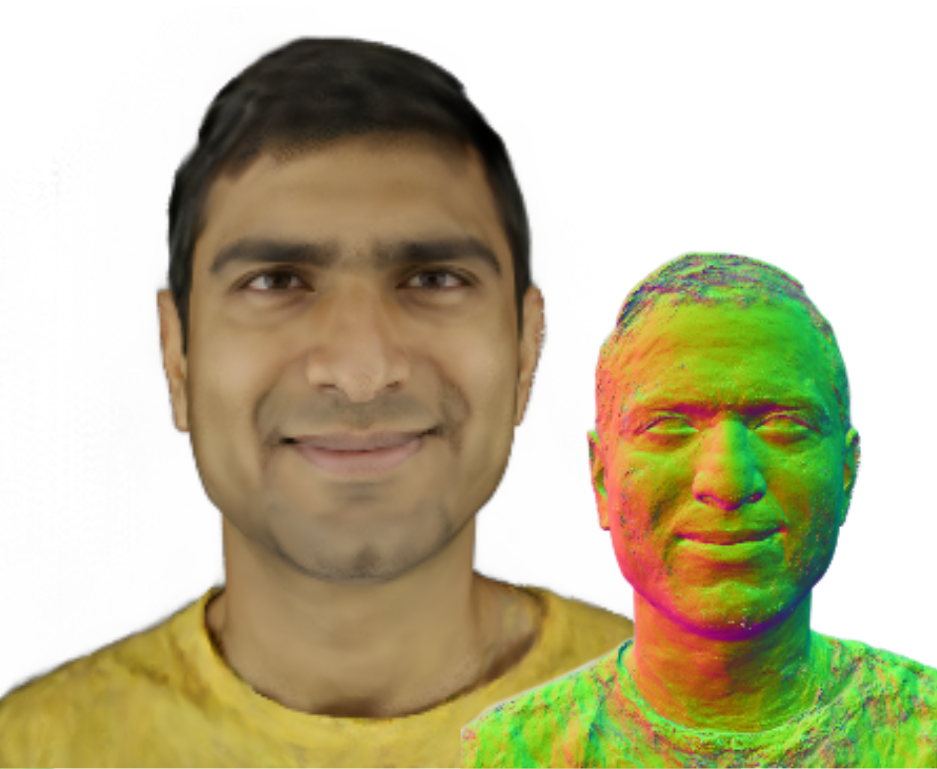} &
    \includegraphics[height=\height]{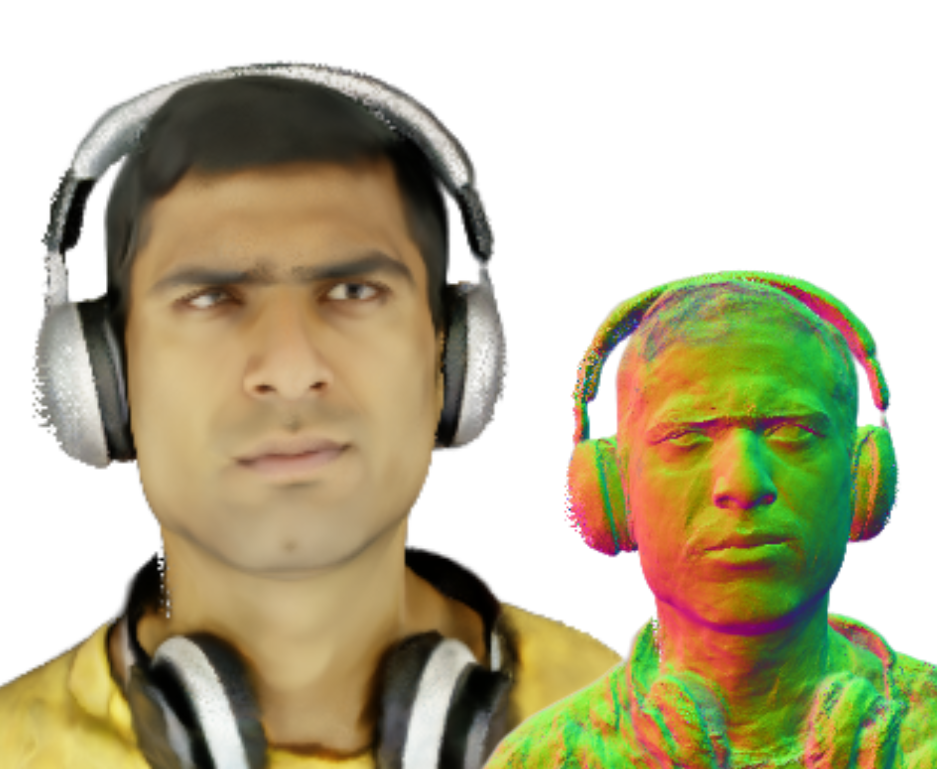} &
    \includegraphics[height=\height]{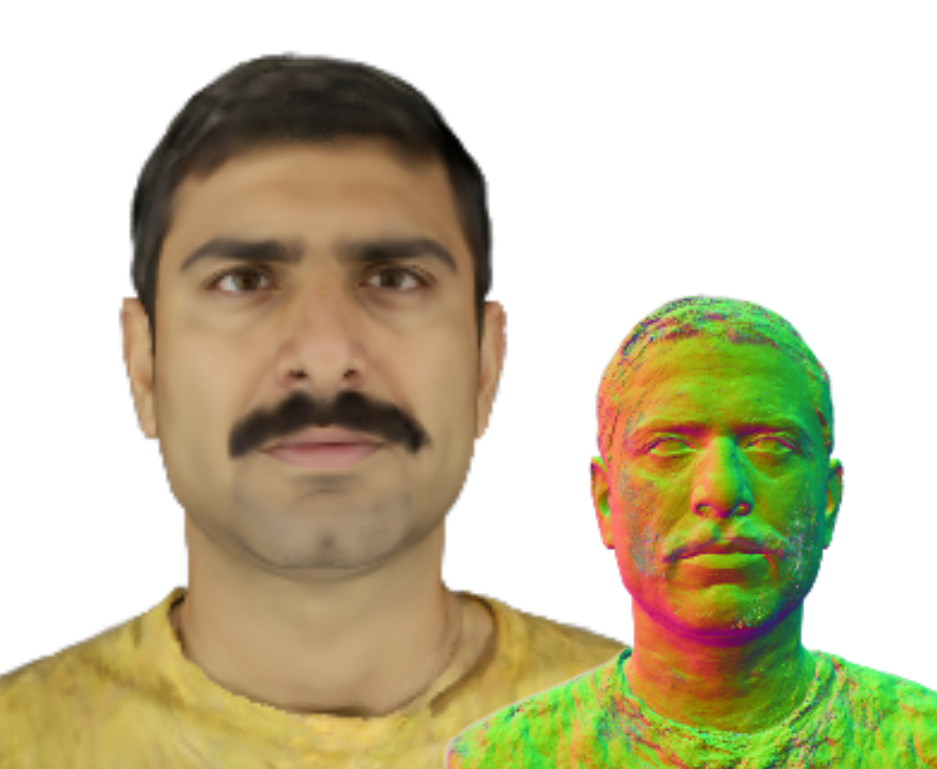} &
    \includegraphics[height=\height]{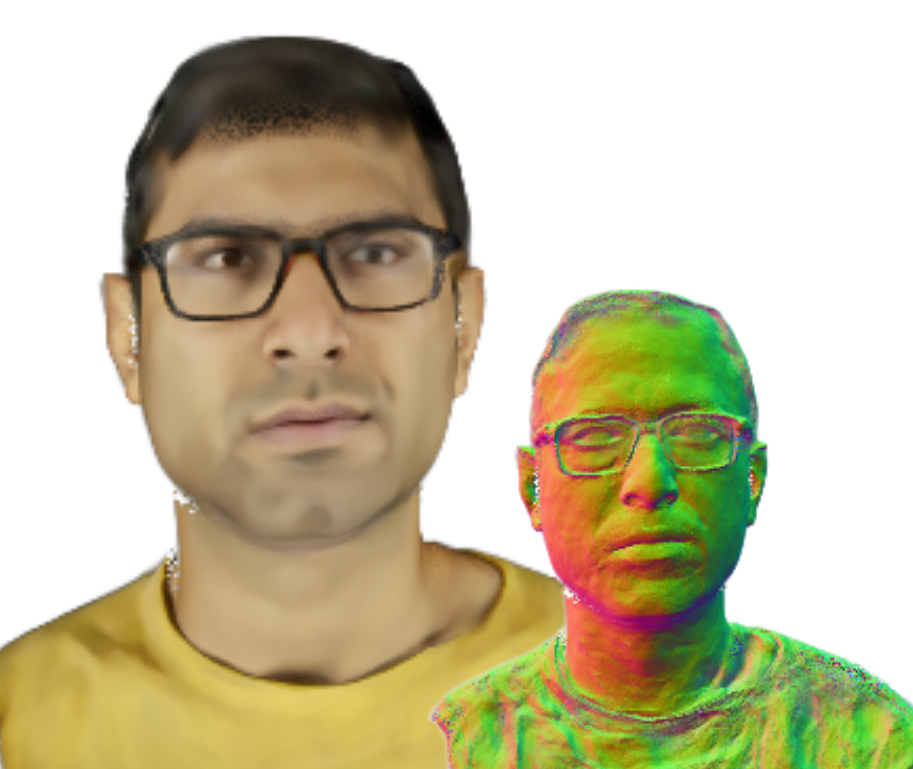} &
    \includegraphics[height=\height]{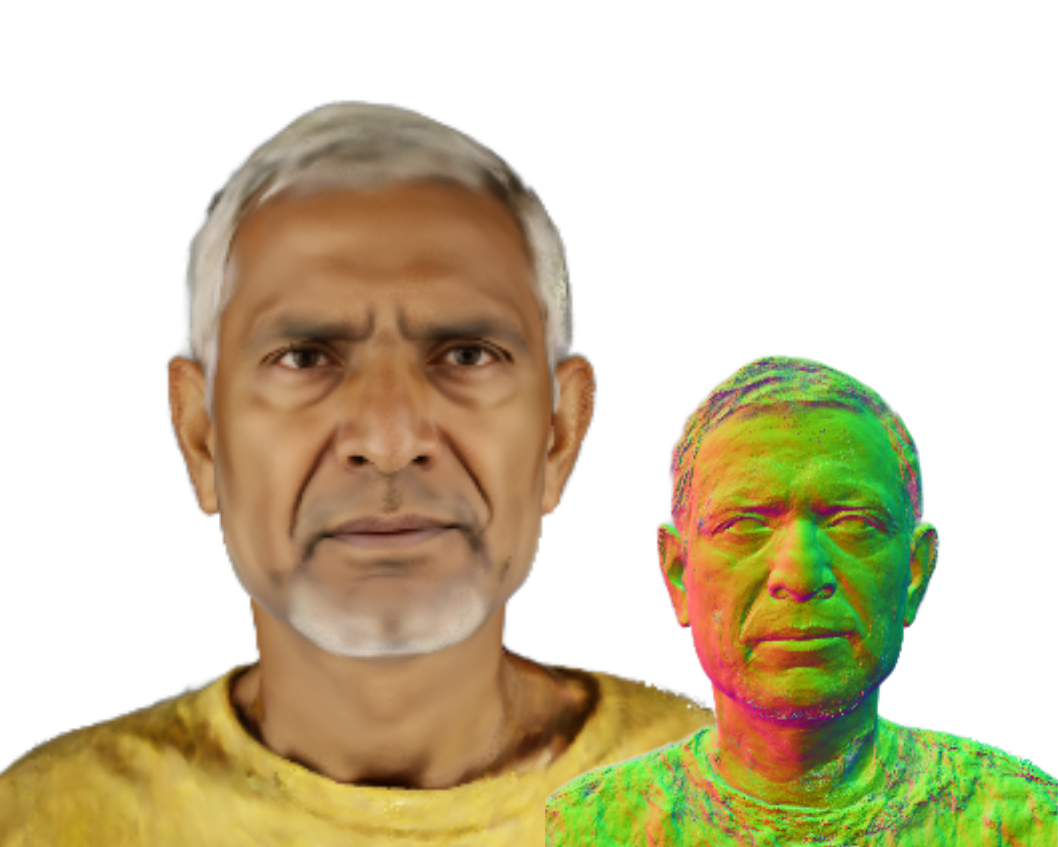} \\
\includegraphics[height=\height]{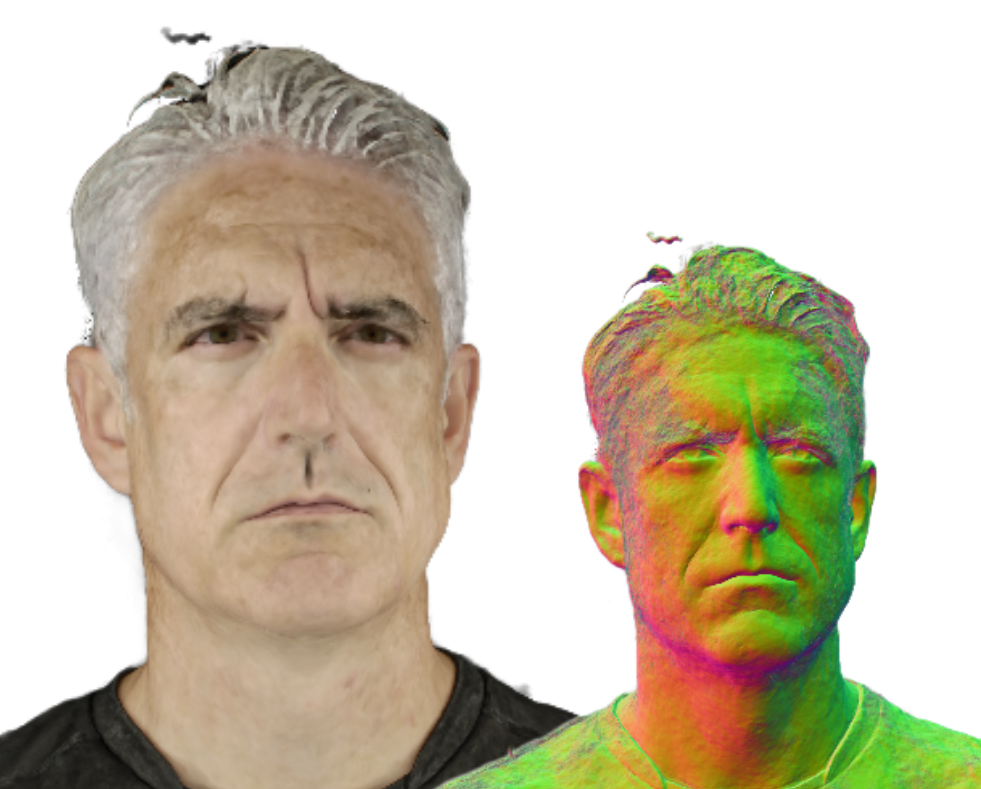} &
    \includegraphics[height=\height]{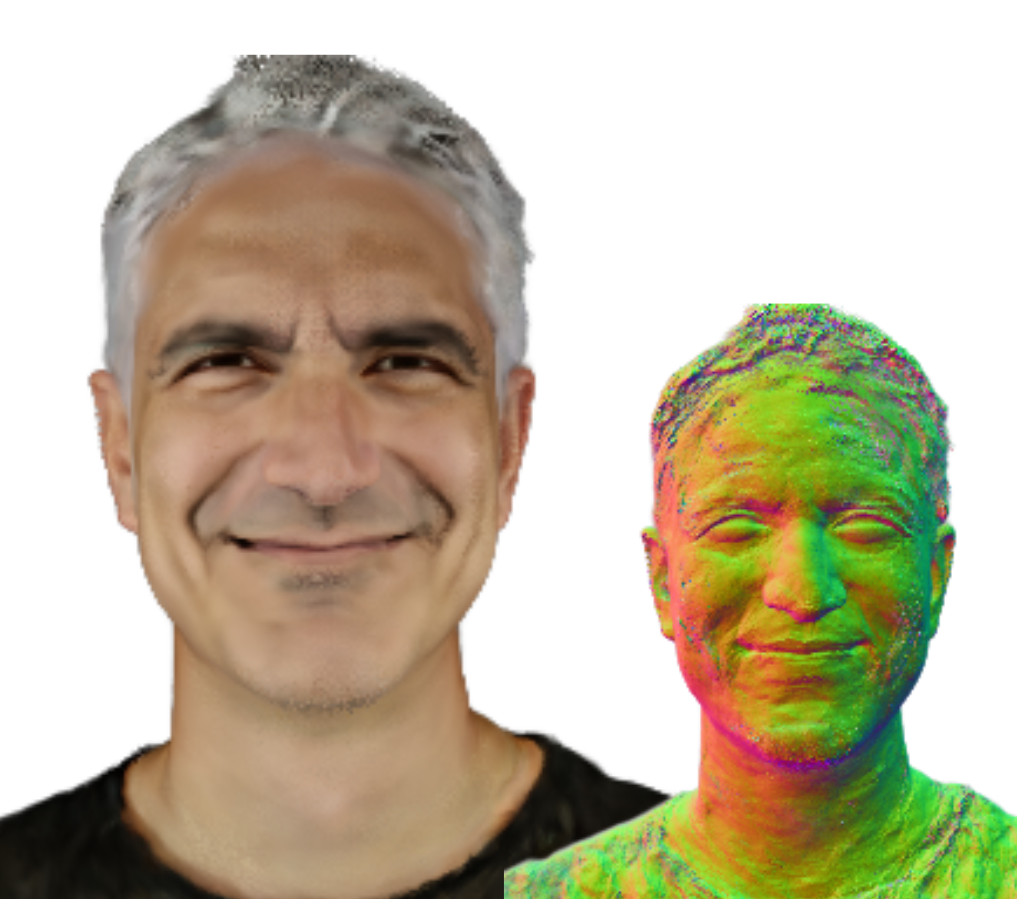} &
    \includegraphics[height=\height]{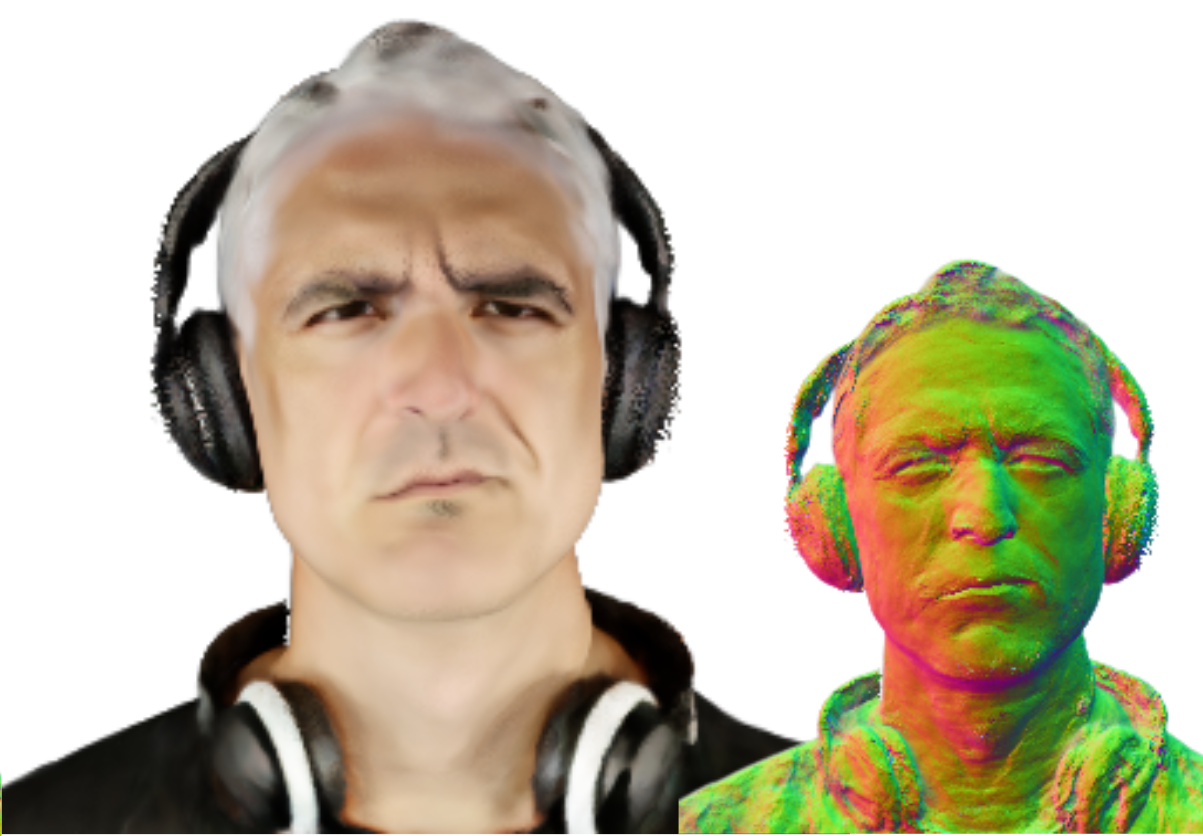} &
    \includegraphics[height=\height]{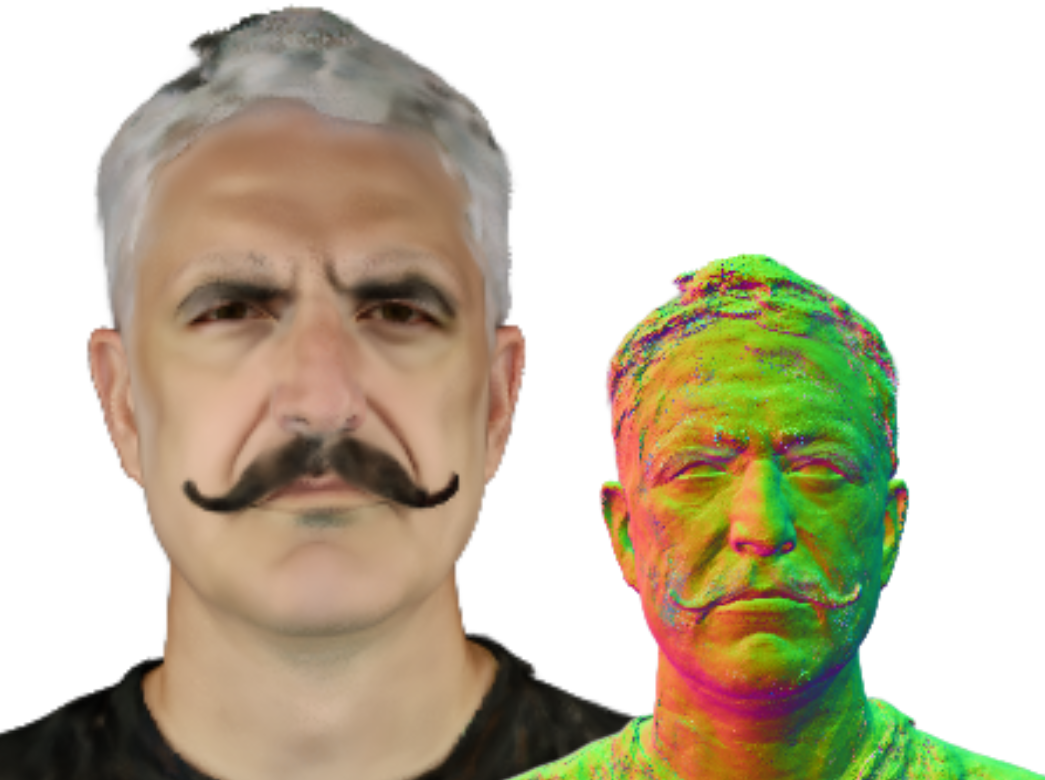} &
    \includegraphics[height=\height]{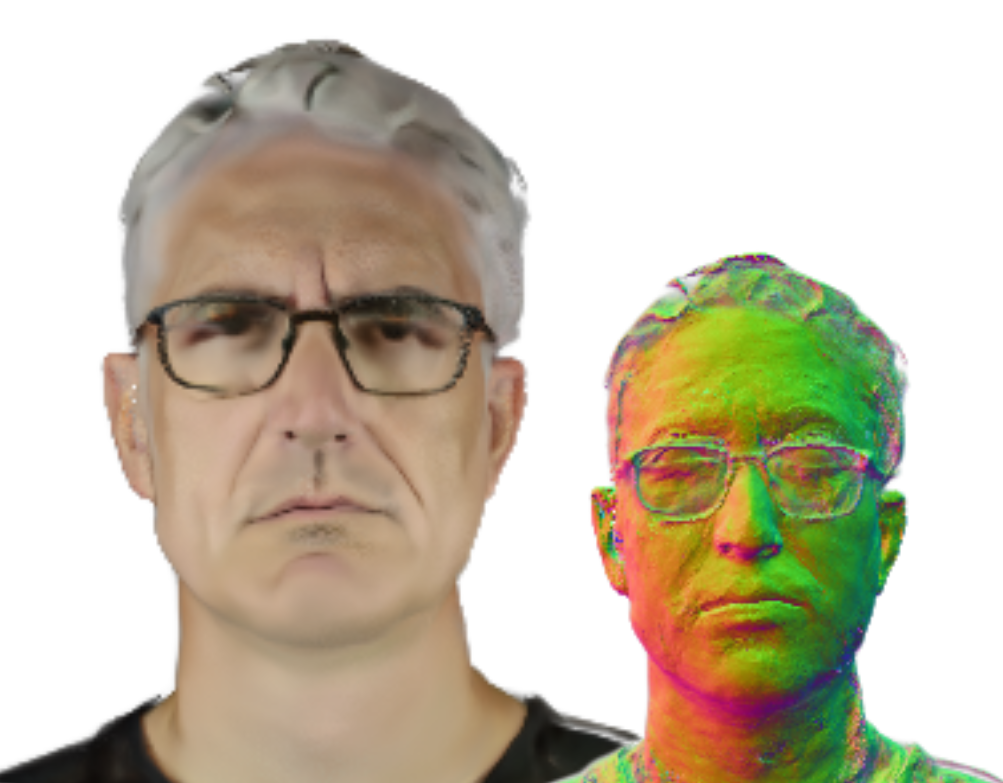} &
    \includegraphics[height=\height]{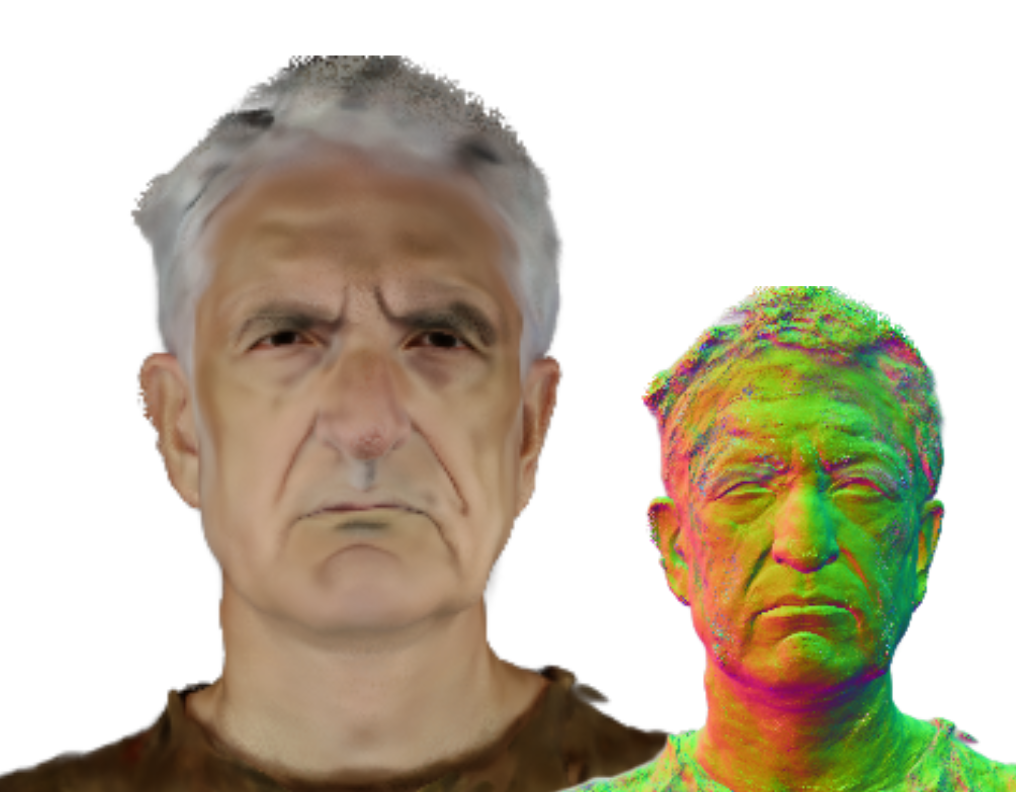} \\
\tiny Original Avatar &\tiny Who is happy & \tiny Wearing headphones &\tiny With a moustache &\tiny Wearing glasses & \tiny Old Person
\end{tabular}
\end{center}
\vspace{-10pt}
\caption{\label{fig:editing}Our novel framework, \model, can successfully change facial expressions, features, and add accessories or specific styles to the person. 
}
\end{figure*}

Another approach is to constrain the space of 3D objects of interest and their representation, similar to popular parametric blendshapes models that enable shape reconstruction given only partial information such as monocular landmarks \cite{Danecek2022emoca}.
For text-guided avatar generation and editing, existing works typically employ an object-specific parametric 3D morphable model (3DMM) \cite{blanz1999morphable} as underlying geometry proxy. However, avatar customization remains challenging because it requires creating novel, semantically meaningful geometric structures that need to introduce new out-of-model elements. So far, because of the crucial dependency on multi-view consistent supervision, it is still challenging to obtain high-quality avatar customizations that closely follow their associated text prompts.

This paper presents \model, our novel framework for text-guided 3D head avatar generation and editing whose visual quality improves upon the current state of the art. Our key idea is to derive constraints and priors that make the test-time optimization problem easier, and less dependent on photometric consistency.
\PG{It is not clear what "test-time optimization" means; first briefly summarize the stages in our method and their goals. What are the inputs and outputs?}
This idea is implemented via the following important framework components:
\begin{enumerate}
    \item A constrained {\bf initial solution space} is first learned as a conditional NeRF model trained on an unannotated multi-view dataset of human heads; this flexible model can express a wide range of head appearances and geometries and subsequently facilitates fast avatar generation and editing.
    \item Leveraging a pre-trained text-to-image diffusion model and its ability to learn new concepts, we build a geometric prior by teaching this model to generate normal maps. This additional {\bf geometry prior} encourages better view invariance, direct geometry optimization, and also largely mitigates the photometric inconsistency problem from conventional multi-view supervision. 
    \item When optimizing our conditional NeRF, Score Distillation Sampling (SDS) \cite{poole2023dreamfusion} can lead to artifacts such as lack of texture and over saturation. We overcome these issues by adopting Variational Score Distillation (VSD) \cite{wang2023prolificdreamer}, allowing us to optimize both appearance and geometry with higher quality.
\end{enumerate}
As demonstrated next, our framework generates custom avatars following specific text instructions with a high level of faithfulness and visual detail. As in DreamBooth\cite{ruiz2023dreambooth,raj2023dreambooth3d}, we leverage a re-contextualization technique that allows users to personalize their avatars with ease and high fidelity to their own identity, while making it fun to create and explore.

\section{Related Work}\label{sec:rel-work}

\subsection{3D Representations for Photorealistic Avatars}
The significance of 3D human modeling has spurred thorough exploration into proper avatar representations. Early methods~\cite{weise2011realtime,ichim2015dynamic,cao2016real,garrido2016reconstruction,thies2016face,hu2017avatar,kim2018deep,tewari2019fml,chaudhuri2020personalized,bai2021riggable} adopt explicit geometry and appearance, particularly parametric human prior models~\cite{blanz1999morphable,loper2015smpl}. However, these approaches struggle with limited representation capabilities.

Lately, the rapid progress in volumetric neural rendering like NeRF~\cite{mildenhall2020nerf} and 3DGS~\cite{kerbl20233d} has promoted implicit avatar modeling, owing to its rendering quality and comprehensive representation. Nevertheless, training such a model typically demands substantial multi-view data for a single subject. To enable monocular inputs and facilitate animation, various human priors have been explored.

One approach involves hybrid representations that leverages morphable models, such as NerFACE~\cite{gafni2021dynamic}, RigNeRF~\cite{athar2022rignerf}, IMAvatar~\cite{zheng2022avatar}, and MonoAvatar~\cite{bai2023learning}. While efficient monocular avatar rendering and animation can be achieved, quality is often compromised due to the limitations of explicit models. Another strategy relies on generative human priors, capable of reconstructing high-quality implicit avatars from sparse inputs. For example, PVA~\cite{raj2021pixel}, CodecAvatar~\cite{cao2022authentic}, Live3DPortrait~\cite{trevithick2023real}, and Preface~\cite{buhler2023preface}. In this work, we follow the latter approach and demonstrate that such a prior not only assists with monocular avatar modeling but also text-driven avatar synthesis.

\subsection{Text-Guided Avatar Generation and Editing}
Generative models have enabled identity sampling within the 2D~\cite{karras2021style} and 3D~\cite{chan2022efficient} latent space. Nonetheless, there is a general preference for better controllability. Among various control modalities, such as scribbles~\cite{chen2021deepfaceediting}, semantic attributes~\cite{shen2022interfacegan}, and image references~\cite{zhu2022mind}, text prompts in natural language are more widely accepted for a broad range of tasks.

The emergence of the language-vision model CLIP~\cite{radford2021learning} has made text-guided avatar editing feasible. In 2D, pioneering work such as StyleGAN-NADA~\cite{gal2022stylegan} transfers pre-trained StyleGAN2~\cite{karras2020analyzing} models to the target style domain described by a textual prompt. This capability extends to 3D as well, where CLIP supervision is integrated with explicit~\cite{hong2022avatarclip} or implicit~\cite{wang2023nerf} human models. However, these models often encounter limitations in expressing full 3D complexity, primarily due to the restricted capacity of CLIP in comprehending intricate prompts.

With the recent advancements in 3D-aware diffusion models, diffusion-based text-guided avatar synthesis have garnered increased attention. DreamFace~\cite{zhang2023dreamface} and HeadSculpt~\cite{han2023headsculpt} introduce coarse-to-fine pipelines to enhance identity-awareness and achieve fine-grained text-driven head avatar creation. HumanNorm~\cite{huang2023humannorm} presents an explicit human generation pipeline, employing normal- and depth-adapted diffusion for geometry generation and a normal-aligned diffusion for texture generation. In a similar two-stage pipeline, SEEAvatar~\cite{xu2023seeavatar} and HeadArtist~\cite{han2023headsculpt} evolve geometry generation from a template human prior and represent appearance through neural texture fields. Meanwhile, AvatarBooth~\cite{zeng2023avatarbooth}, AvatarCraft~\cite{jiang2023avatarcraft}, DreamAvatar~\cite{cao2023dreamavatar}, DreamHuman~\cite{kolotouros2023dreamhuman}, and DreamWaltz~\cite{huang2023dreamwaltz} propose text-driven avatar creation utilizing implicit surface representation, parameterized with morphable models for easy animation. In terms of editing, AvatarStudio~\cite{mendiratta2023avatarstudio} achieves personalized NeRF-based avatar stylization through view-and-time-aware SDS on dynamic multi-view inputs.

\subsection{3D-Aware Diffusion Models}
The success of text-to-image diffusion models~\cite{rombach2022high} naturally encourages researchers to explore 3D-aware diffusion. Building from 2D diffusion, many studies have concentrated on synthesizing consistent novel 2D views of 3D objects, such as 3DiM~\cite{watson2023novel}, SparseFusion~\cite{zhou2023sparsefusion}, and GeNVS~\cite{chan2023generative}. Zero-1-to-3~\cite{liu2023zero} proposes a pipeline that fine-tunes a pre-trained diffusion model with a large-scale synthetic 3D dataset. SyncDreamer~\cite{liu2023syncdreamer} further improves the cross-view consistency.

Direct 3D generation have also been explored across various representations, including point clouds~\cite{luo2021diffusion,nichol2022point}, feature grids~\cite{karnewar2023holodiffusion}, tri-planes~\cite{shue20233d,wang2023rodin}, and radiance fields~\cite{jun2023shap}. However, due to the complexity of representations, heavy architecture, and shortage of large-scale 3D data, 3D diffusion often suffers from poor generalization and low quality results.

Compared to 3D diffusion models, lifting 2D diffusion for 3D generation is more appealing, spearheaded by the pioneering works of DreamFusion~\cite{poole2023dreamfusion} and SJC~\cite{wang2023score}. At the heart of these approaches lies score distillation sampling (SDS), which employs 2D diffusion models as score functions on sampled renderings, providing supervision for optimizing the underlying 3D representations. Subsequent works like DreamTime~\cite{huang2023dreamtime}, MVDream~\cite{shi2023mvdream}, and ProlificDreamer~\cite{wang2023prolificdreamer} refine the architectural design with better sampling strategy, loss design, and multi-view prior. Meanwhile, Magic3D~\cite{lin2023magic3d}, TextMesh~\cite{tsalicoglou2023textmesh}, Make-It-3D~\cite{tang2023make}, and Fantasia3D~\cite{chen2023fantasia3d} extend the approach to other representations such as textured meshes and point clouds. Notably, variational score distillation (VSD) is proposed in ProlificDreamer to address oversaturation and texture-less issues of SDS. Our method also adopts VSD to enhance the quality of the generated results.

\section{Method}\label{sec:method}
We present two similar pipelines for (P1) text-driven generation and (P2) personalized 3D head avatars editing. Both pipelines have the same structure, illustrated in Fig.~\ref{fig:arch}. We render random views from our avatar, chosen randomly from a set of orbit renders. The avatar is parameterized by a conditional NeRF model and initialized with any latent identity code (Sec.~\ref{sec:initialization}). We then employ a distillation approach with a \textit{geometry prior} to optimize the initial avatar's NeRF appearance and density, following the methodology described in Section \ref{sec: objective}.
In both pipelines, the geometry is captured by a diffusion prior, which is fine-tuned to capture facial geometry features from a single avatar (Sec. \ref{sec:geoprior}).

More specifically, besides the conditional NeRF, in Pipeline P1 our method mainly leverages a pre-trained text-to-image diffusion models that captures the distribution of real RGB images. The diffusion model and the geometric prior allows us to customize both the appearance and geometry of our initial NeRF avatar as guided by an input text prompt in the form: \emph{``A portrait of a [source description]''}.  There is no personalization element in P1. Thus, the prompt does not require a subject identifier. Avatar customization is done using a distillation-based objective function derived from Variational Score Distillation (VSD)~\cite{wang2023prolificdreamer} (Sec.~\ref{sec: objective}). 

In Pipeline P2, we personalize on a particular subject by first conditioning on user-provided 2D images in multiple views, or by rendering images from a reconstructed digital asset of the target subject. This subject is associated with a unique identity token {\em [V]} and {\em [source description]}, using DreamBooth to fine-tune our text-to-image diffusion model with the text prompt {\em "A [V] portrait of [source description]"}. 

The user can then supply a new {\em [target description]} prompt to guide avatar stylization to their preference, which is achieved by optimizing the conditional NeRF using the objective defined in \eqref{eq: final objective}, with the text embedding corresponding to {\em "A [V] portrait of [target description]"}. 

Finally, a user can combine multiple priors in parallel to achieve different objectives. Updates from subject-aware texture priors can be blended with updates from text-to-image generic diffusion priors, prompted with multiple context prompts.

\subsection{Constraining the solution space} \label{sec:initialization}

To lift the partial 2D information from text-to-image diffusion models, a 3D prior model is needed to constrain the optimization domain.
We thus parameterize the subspace of 3D head avatars using a conditional NeRF, which simplifies the optimization while remaining flexible enough to accommodate variations outside the training data.

To learn our solution subspace, we leverage Preface~\cite{buhler2023preface}, a conditional model that extends Mip-NeRF360~\cite{mipnerf360} with a conditioning (identity) latent code concatenated to the inputs of each MLP layer.
Here, we briefly describe how to train this conditional NeRF on our multiview dataset of $1450$ human faces with a neutral expression, captured by 13 synchronized cameras under uniform in-studio lighting. More details can be found in the Supplementary material.
Each face is assigned a learnable latent code \cite{bojanowski2018optimizing} that is optimized together with the model weights, under the supervision of pixelwise reconstruction loss only.
Each training batch randomly samples pixel rays from all subjects and cameras, promoting generalization over the space of human faces rather than over-fitting to a few subjects.
The importance of the diversity of training faces is highlighted in our ablation study in Sec.~\ref{sec:num_subjects}. 

\subsection{View-invariant geometric prior}
\label{sec:geoprior}
\setlength{\belowcaptionskip}{-2pt} 
\begin{wrapfigure}{l}{0.4\textwidth}
\vspace{-1cm}
  \includegraphics[width=0.38\textwidth]{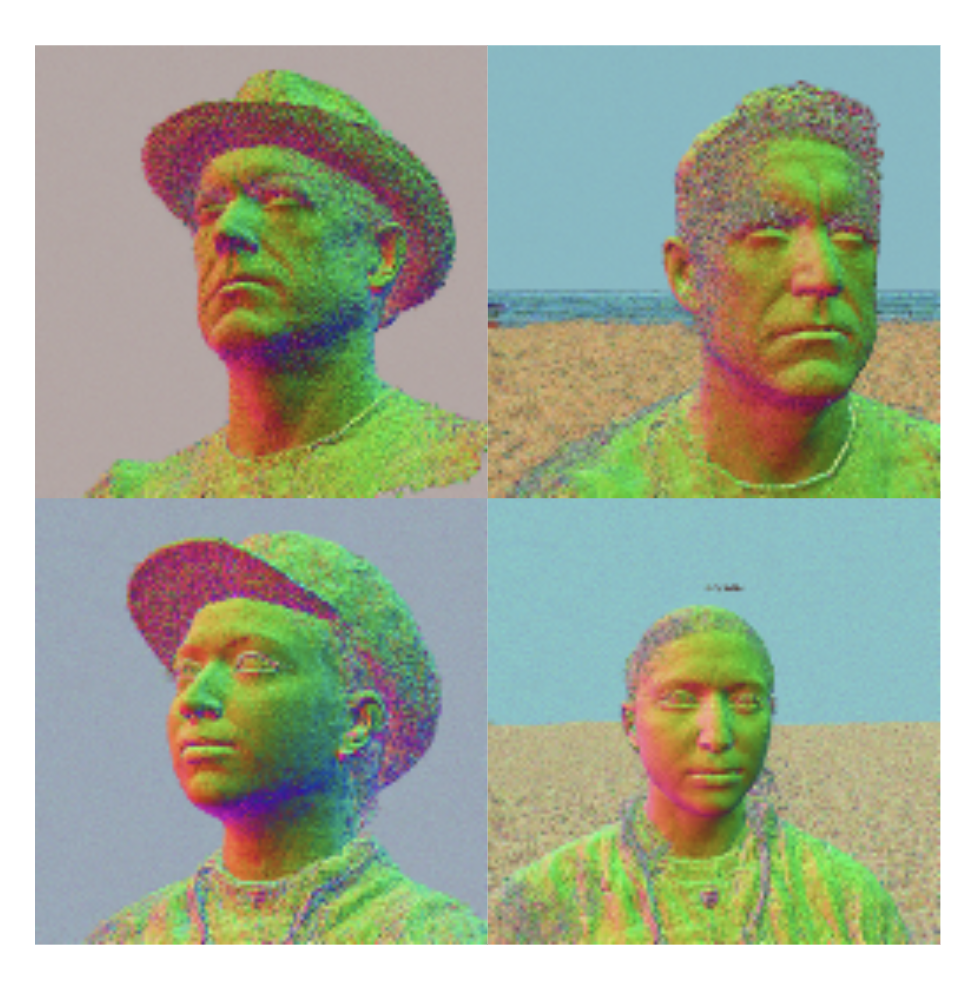}
  \vspace{-0.2cm}
  \caption{Text-to-Image Diffusion Model have the remarkable ability to re-contextualize new concepts. We show the generated normal maps under new text prompts. Note that they are {\it not} rendered from a NeRF.}
  \label{fig:normals}
  \vspace{-1.1cm}
\end{wrapfigure}

Given the above conditional NeRF that models diverse 3D head avatars, we now turn to incorporating high-frequency geometric details that are (1) authentic to specific target user and (2) consistent with a given text prompt.
To this end, we show that an additional pretrained text-to-image diffusion model $\mathcal{D}_{\text{color}}$ can be used to capture new appearance and geometry concepts without any architectural change. Instead of retraining the diffusion model to encourage multiview consistency, we propose a novel and effective solution that teaches the model to also generate normal maps of human heads, effectively deriving a second model $\mathcal{D}_{\text{normal}}$.

Our solution leverages the few-shot learning technique proposed by DreamBooth~\cite{ruiz2023dreambooth}: given $60$ world-space surface normal renderings from different camera views of an avatar, we pair them with text descriptions such as {\em "A [W] face map of a man/woman [source description]"}, where {\em "[W]"} is the unique identifier for the new concept of {\em surface normal}.

Next, we fine-tune the text-to-image diffusion model using these text-annotated surface normal renderings. As a result, the fine-tuned diffusion model $\mathcal{D}_{\text{normal}}$ can now also predict reasonable surface normal maps for new heads, even when re-contextualized to the rest of the text prompt description (Fig.~\ref{fig:normals}). This new capability provides additional geometric critique for avatar generation and editing, driving the optimization to go beyond simple color edits to effectively improve the geometry.

\setlength{\belowcaptionskip}{0pt} 
When defining the input data for fine-tuning the diffusion model, it is crucial that surface normals be defined in a fixed world coordinate system that is aligned with our solution space, hence independent of camera viewpoint.
The source normal maps can then be obtained from any subject, as evidenced in Sec.~\ref{sec:ablation normals}.

\subsection{Test-time optimization objective} \label{sec: objective}

During our test-time optimization, the fine-tuned diffusion model is re-contextualized with the prompt {\em "A [V] portrait of a [target description]"}. Such geometric prior is then incorporated via a Variational Score Distillation (VSD)~\cite{wang2023prolificdreamer} optimization objective, as described in the following.

To derive our solution, we first review Score Distillation Sampling (SDS)~\cite{poole2023dreamfusion}, which minimizes the following loss function:
\begin{equation}
    \mathcal{L}_{\text{SDS}}(\text{sg}(\mathcal{D}), I, \epsilon, T, t) = \omega(t)\| \text{sg}(\mathcal{D}(I, \epsilon, T, t)) - I \|^2,
\end{equation}
where $\mathcal{D}$ represents the text-to-image diffusion model that outputs the denoised image by processing the NeRF rendering $I$, Gaussian noise $\epsilon$, a fixed target text embedding $T$, and a time parameter $t$ that follows certain annealing schedule $t \to 0$.
$\text{sg}(\cdot)$ denotes the stop gradient operator,
and $\omega(t)$ is a time-dependant weighting factor.
In what follows, some or all of the loss arguments may be omitted when the context is clear.

It is commonly observed that SDS, with its default high Classifier-Free Guidance (CFG) weight~\cite{ho2022classifier}, often leads to textureless and over-saturated outputs that significantly impact photorealism and diversity, while a lower CFG weight tends to underperform with SDS.

Variational Score Distillation (VSD)~\cite{wang2023prolificdreamer} introduces a proxy of $\mathcal{D}$, denoted as $\mathcal{D}'$, which is optimized under the following loss function:
\begin{equation}
    \mathcal{L}_{\text{proxy}}(\mathcal{D}', \text{sg}(I)) = \omega(t)\| \mathcal{D}'(I, \epsilon, T, t) - \text{sg}(I) \|^2. 
\end{equation}
Typically, $\mathcal{D}'$ is selected to be the Low Rank Adaptation (LoRA)~\cite{hu2022lora} of $\mathcal{D}$, having identical outputs to $\mathcal{D}$ at the beginning of the optimization. Simultaneously, VSD also optimizes the NeRF parameters by minimizing $\mathcal{L}_{\text{SDS}}(I) - \mathcal{L}_{\text{proxy}}(\text{sg}(\mathcal{D}'), I)$.
The full VSD objective is formulated as:
\begin{equation}
    \mathcal{L}_{\text{VSD}}(\mathcal{D}', I) = \mathcal{L}_{\text{SDS}}(I) - \mathcal{L}_{\text{proxy}}(\text{sg}(\mathcal{D}'), I) + \mathcal{L}_{\text{proxy}}(\mathcal{D}', \text{sg}(I)).
\end{equation}

Leveraging the formulation above, our overall test-time optimization objective is:
\begin{equation} \label{eq: final objective}
    \mathcal{L}_{\text{ours}} =
    \mathcal{L}_{\text{VSD}}(\mathcal{D}_{\text{color}}', I_{\text{color}}) +\lambda \mathcal{L}_{\text{VSD}}(\mathcal{D}_{\text{normal}}', I_{\text{normal}}),
\end{equation}
where two fixed text-to-image diffusion models, $\mathcal{D}_{\text{color}}$ and $\mathcal{D}_{\text{normal}}$, process color images $I_{\text{color}}$ and normal maps $I_{\text{normal}}$, respectively.
The normal map is computed through analytic gradients of the NeRF density.
During VSD, we also optimize their LoRAs, $\mathcal{D}_{\text{color}}'$ and $\mathcal{D}_{\text{normal}}'$.
Our observations suggest that VSD allows for a smaller CFG weight, generally enhancing convergence and output quality.

\paragraph{Mixing and weighting concepts:} Through this distillation approach, we can additionally compose and modulate different concepts expressed by the text embedding. 
We proposevv to perform compositionality using a perspective akin to findings in the context of Energy-Based Models \cite{Du2020CompositionalVG, Liu2021LearningTC} and Diffusion Models \cite{Liu2022CompositionalVG}. 
This notion allows us to generate a variety of results by mixing and weighting two or more concepts, including removal of certain concepts or semantic interpolation, thereby enriching the user experience.
More generally, we can optimize the conditional NeRF from initialization with a combination of objectives:
\begin{equation} 
    \mathcal{L}_{\text{composed}} = \sum_{T \in \text{Positive}} \alpha_{T} \cdot \mathcal{L}_{\text{ours}}(T) - \sum_{T \in \text{Negative}} \beta_{T} \cdot \mathcal{L}_{\text{ours}}(T)
\end{equation}
with $\{\alpha_T, \beta_T\}$ being the positive modulation constants, balancing the importance of different concepts. A probabilistic interpretation is provided in the supplementary material.
The associated updates to the NeRF paramenters $\theta$ are thus
\begin{equation}
\label{eq:vsd-multiconcept}
\begin{aligned}
\nabla_{\theta}\mathcal{L}_{\text{composed}}(T, \theta) =  \sum_{T \in \text{Positive}} \alpha_{T} \cdot \nabla_{\theta}\mathcal{L}_{\text{ours}}(T, \theta) - \sum_{T \in \text{Negative}} \beta_{T} \cdot \nabla_{\theta}\mathcal{L}_{\text{ours}}(T, \theta)
\end{aligned}
\end{equation}

which is an expression reminiscent of the concept-compositional sampling by means of Langevin Dynamics in Energy-Based Models~\cite{Du2020CompositionalVG}.

Besides the composability interpretation, we can generate smooth interpolations by simply switching from one concept to another.
This is, once we already obtained the result from one concept, we can directly apply the optimization with the objective associated to an alternate concept.
We observe that the optimization trajectory tends to remain within distribution, assuming the two concepts do not introduce significant changes, such as the opening of the mouth or the addition of extra geometry from accessories.
We illustrate these interesting findings in Sec.~\ref{sec:exp} and hypothesize that incorporating additional data into training our 3D prior model could lead to more meaningful optimization trajectories. Experiments on both methodologies can be found in Sec.~\ref{sec:exp compositionality}.

\subsection{Implementation details}

For a single prompt, \model needs a maximum of 1k iterations for both geometry and texture generation, which are performed simultaneously. We utilize 4 TPUs of 96 GB of memory, with a batch sample of $128 \times 128$ resolution per device. Each device may leverage a different set of weights for its diffusion prior, all of them implemented with Imagen 2.2.3. The entire generation process takes about 15 minutes. Additional details can be found in the supplementary.

\subsection{Ablation studies}

\subsubsection{Role of (personalized) geometric prior.} \label{sec:ablation normals}
One noteworthy property of the geometry supervision is its robustness. Fig.~\ref{fig: geo prior ablation} (top row) illustrates the crucial role played by the geometry prior. Without it, the normal map appears noisy and distorted, with out-of-face structures like headphones poorly constructed. Figure \ref{fig: geo prior ablation} (bottom row) shows that despite our geometry prior being trained on a single avatar, its identity has a negligible impact on the final results. In this comparison, we analyze the effects of a geometry prior trained on the original normals of the subject (shown on the right) against one trained on normals from a random female (shown on the left). The results are visually indistinguishable, with both facial features and overall geometry are accurately generated.

\subsubsection{Test-time NeRF initializations.}
We test different latent codes to initialize our conditional NeRF for test-time avatar optimization.
As can be seen in \ref{fig: nerf init ablation}, for personalized avatars, different latent codes does not yield significant variance in the final results.
The non-personalized avatars can have more variance since there are no more constraints on the appearance than to follow the text prompt.

\subsubsection{Diversity of training data for conditional NeRF.} 
\label{sec:num_subjects}
The performance of \model significantly benefits from the diversity of the search space. As aforementioned, employing a conditional NeRF trained on multi-view datasets comprising multiple subjects has proven to be valuable. As illustrated in Figure \ref{fig: nerf data ablation}, the final results are notably influenced by the number of subjects used for training the conditional NeRF. Training with a single subject tends to yield very rigid geometry, much more challenging to modify compared to the texture. Utilizing $350$ subjects allow for modification in geometry but often results in rough normals and a lack of fine details. Conversely, a more diverse training set of $1450$ subjects leads to substantially smoother and more precise geometry. We choose the complete set for all of other experiments.

\subsubsection{Impact of the avatar's identity latent code}
In Fig.~\ref{fig:sds ablation}, we demonstrate the results of optimizing the conditional latent code in isolation. In this experiment, we perform VSD while freezing the remaining parameters of the network. According to the results, while substantial modifications to the overall geometry and texture are achievable through NeRF conditioning, the solution space is inherently limited to human-like faces, as dictated by the training set. The Figure illustrates how latent inversion fails to accurately capture the green color and distinct facial features of a fantasy character "the Grinch".

\subsubsection{VSD vs. SDS.}
\begin{figure*}[t!]
    \centering
        \begin{subfigure}[t]{0.46\linewidth}
            \includegraphics[width=\textwidth]{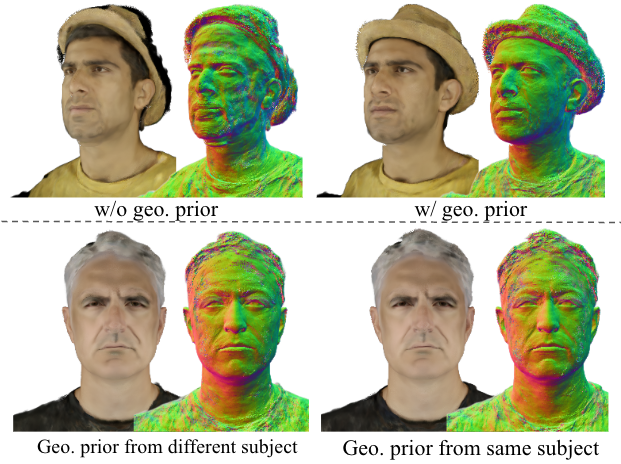}
            \caption{Role of (personalized) geometric prior}
            \label{fig: geo prior ablation}
        \end{subfigure}
        \begin{subfigure}[t]{0.52\linewidth}
            \includegraphics[width=\textwidth]{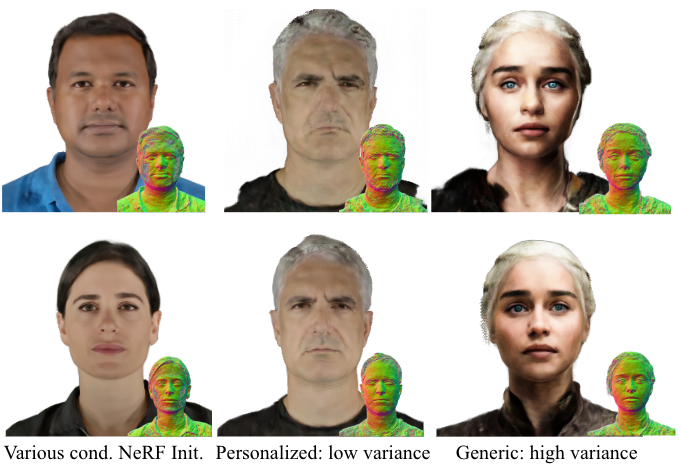}
            \caption{Test-time NeRF initializations}
            \label{fig: nerf init ablation}
        \end{subfigure}
    \begin{subfigure}[t]{.97\linewidth}
        \includegraphics[width=\textwidth]{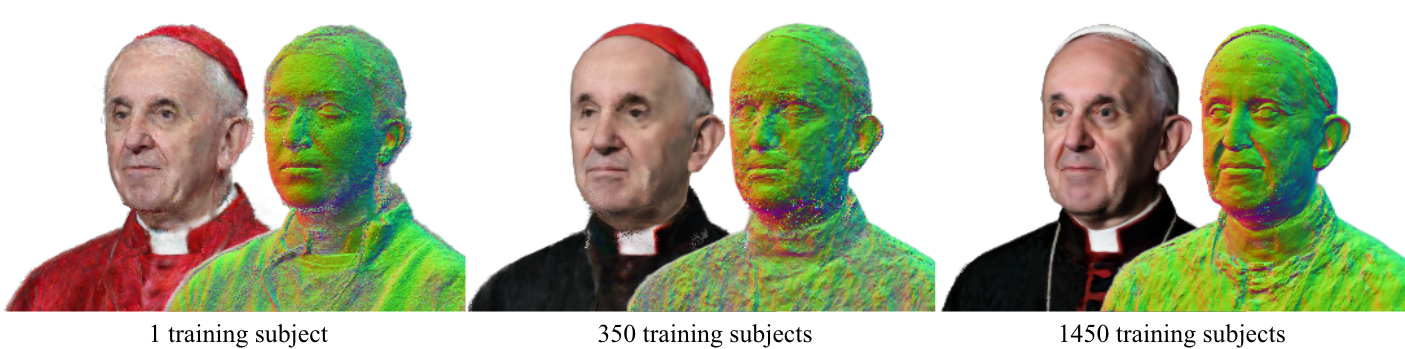}
        \caption{Diversity required for training cond. NeRF}
        \label{fig: nerf data ablation}
    \end{subfigure}
        \centering
        \begin{subfigure}[t]{.53\linewidth}
            \includegraphics[width=\textwidth]{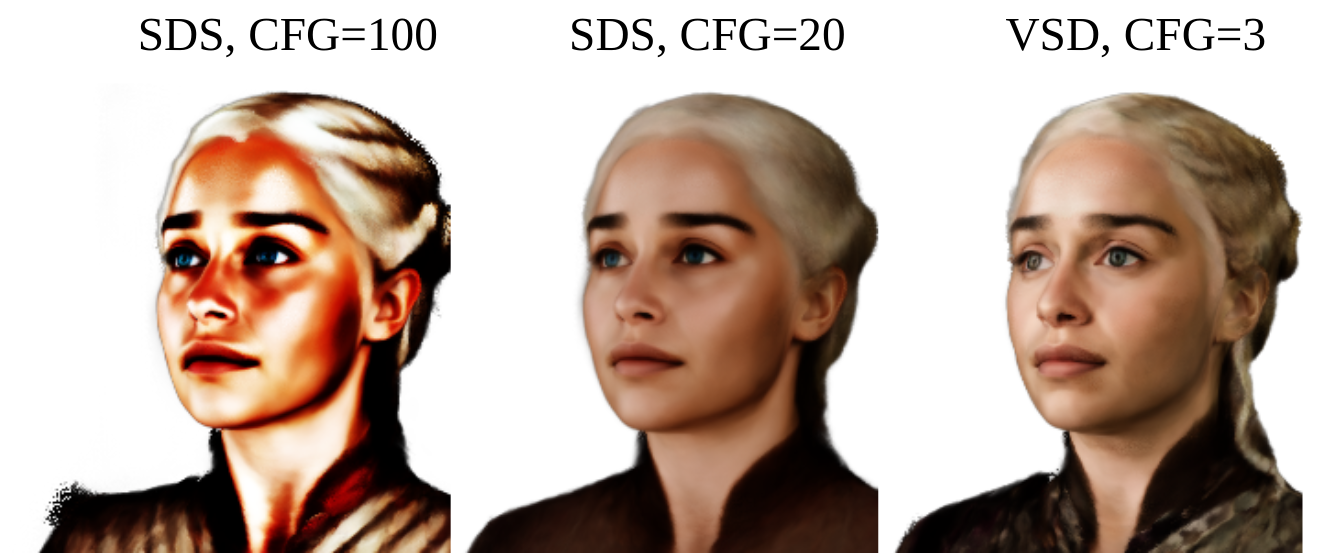}
            \caption{SDS w/ high CFG causes over-smoothing}
            \label{fig:sds ablation}
        \end{subfigure}
        \begin{subfigure}[t]{.44\linewidth}
            \includegraphics[width=\textwidth]{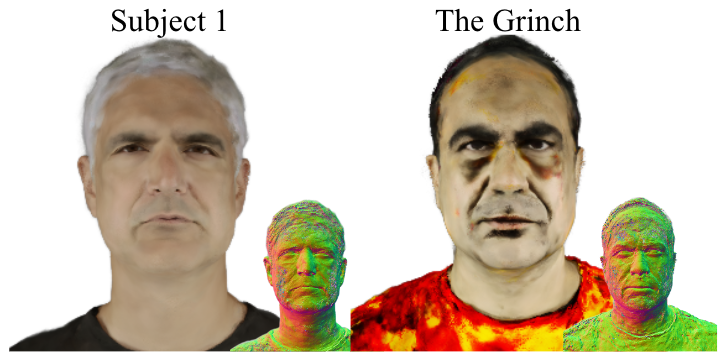}
            \caption{Latent inversion fails out-of-distribution}
            \label{fig:latent inversion ablation}
        \end{subfigure} 

    \caption{Ablation studies. We show that a geometric prior (\ref{fig: geo prior ablation}) improves the results (top-left) even when the geometry prior comes from a different subject (center-left). (top-right) Our method yields very similar results in the personalized setting, even for very different NeRF initializations (\ref{fig: nerf init ablation}). A sufficiently diverse prior (\ref{fig: nerf data ablation}) is required for convincing results (middle). (bottom-left) we demonstrate the effectivenes of VSD instead of SDS. (bottom-right) we show how inverting the latents works to a certain extent but it fails for out-of distribution cases.}
\end{figure*}

In this final assessment, we evaluate the advantages of VSD over SDS regarding avatar generation in our system. Fig.~\ref{fig:latent inversion ablation} presents results for both a real captured individual and a fictional character, employing CFG weights of 20 and 100. These comparisons lead us to observe that SDS tends to produce avatars with an overly smooth and saturated appearance that is deficient in fine detail, as extensively documented in existing literature. In contrast, our implementation of VSD yields avatars that exhibit significantly improved realism and finer details, demonstrating the superiority of VSD in our method.

\begin{figure*}[ht]
\begin{center}
\small
\includegraphics[width=\textwidth]{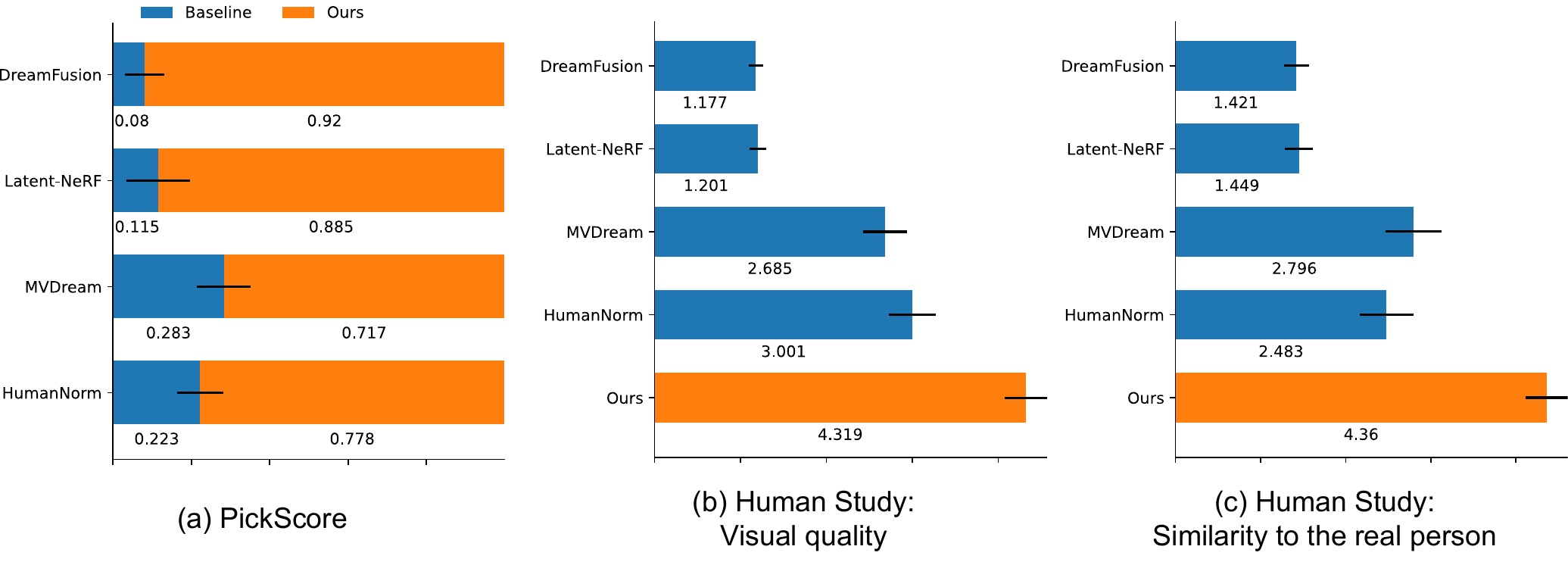}
\end{center}
\vspace{-10pt}
\caption{\label{fig:pickscore} Quantitative evaluations of our method. (a) We compute PickScore~\cite{Kirstain2023PickaPicAO} against each baseline. The Bars indicate the percentage of times that our avatar is preferred w.r.t. the baselines. (bc) We report average human study scores regarding visual quality (b) and similarity to the real person (c) of all baselines. Scores range 1 to 5.
}
\end{figure*}

\section{Experiments}\label{sec:exp}
\subsection{Metrics and Evaluation}
It is well known that evaluation is challenging when it comes to 3D generation. Hence, we resort to human preference to assess the quality of our model. We create a human study, where we show 3 rendered views of 18 generated recognizable subjects for all (anonymous) methods to 36 people and ask them to rank them from 1 (low) to 5 (high) in two dimensions. We ask about (i) \textit{visual quality} as a measure of realism of both shape and appearance of the generated avatars as a generic human being; and (ii) \textit{similarity to the real person} as a measure of alignment to the real target person's identity.
Finally, we utilize the same set of views to provide results for PickScore \cite{Kirstain2023PickaPicAO}, which quantifies human preference.

\subsection{Quantitative Results}
We observe in Fig. \ref{fig:pickscore}(bc) that \model outperforms baselines by a large margin ($>1.5$ compared to the best baseline), achieving very high marks on both questions. 

In addition, Fig. \ref{fig:pickscore}(a) illustrates a similar trend as seen in the human evaluation. Our method is chosen over baselines the majority of times.

\subsection{Qualitative Results}

\subsubsection{Mixture of concepts} \label{sec:exp compositionality}
\begin{figure*}[t!]
\begin{center}
\begin{subfigure}{\textwidth}
    \centering
    \includegraphics[width=0.99\textwidth]{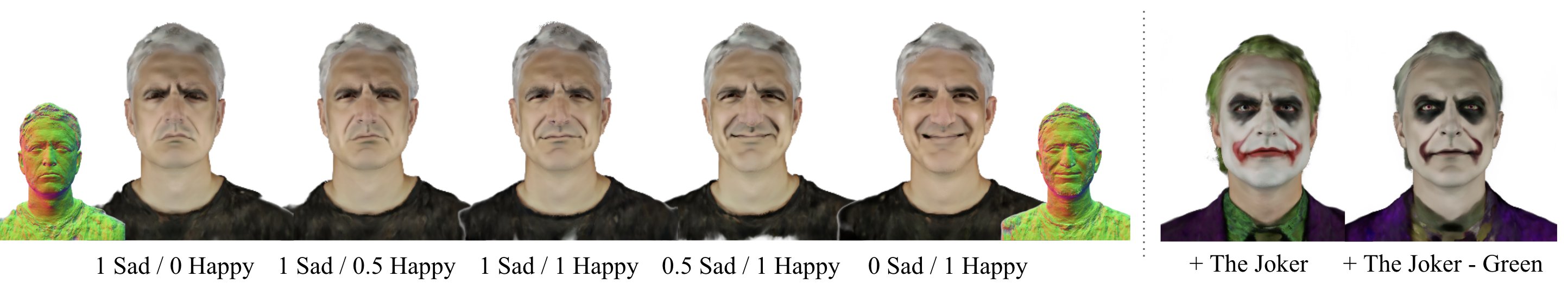}
    \caption{The final optimized results under various modulation weights for two different concepts. Removing a certain visual elements is also possible with this method, for example removing the green color from the Joker re-contextualization for the given identity.}
    \label{fig: moc general}
\end{subfigure}
\begin{subfigure}{\textwidth}
    \centering
    \includegraphics[width=0.9\textwidth]{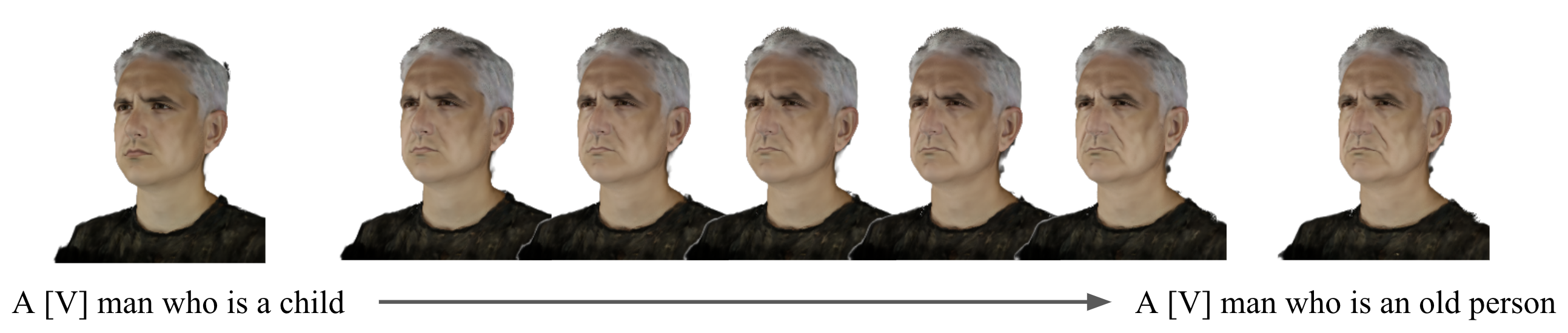}
    \caption{The optimization trajectory from one concept to  another. This works well when there are no drastic change in geometry.}
    \label{fig: moc trajectory}
\end{subfigure}
\caption{Applying our editing framework with mixture of concepts.}
\label{fig: moc}
\end{center}
\end{figure*}

We illustrate the mixture of concepts technique through mixture of objectives described in Sec. \ref{sec: objective}.
In Fig. \ref{fig: moc general} we show the optimized NeRF under various modulation weights for the "happy" and "sad" concepts, as well as an example of removing "green" from "joker".
All NeRF are initialized from the common initialization.
We can see the mixture results in natural and plausible appearance while retaining the same quality of a single concept.
In Fig. \ref{fig: moc trajectory} we show that it's also possible to move from the concept "young" to "old" by optimizing the target objective based on the optimized source objective, although the trajectory may not always make sense if the intermediate states are too out-of-distribution.
Recall that our conditional NeRF is only trained on neutral expression.

\subsubsection{Identity preserving editing with text prompt} \label{sec:exp editing from photos}

AvatarStudio \cite{mendiratta2023avatarstudio} is a recently proposed text guided avatar editing method.
It aims to modify the appearance and geometry of the 3D avatar using the SDS technique.
The 3D avatar is represented as a conditional NeRF while the conditioning on the time in a expression performance. Thus unlike our approach, there is no modelling of different identities.
We compare our results using the same identity and text prompt.
It's worth mentioning that we do not have to reconstruct the user at test time, since we only require photos to be used by DreamBooth.
Thus unlike AvatarStudio, we don't need the camera pose estimation that may be hard to obtain with user's casually captured photos.
We observe that we achieve significant improvement in both visual detail and realism, largely benefit from our constrained solution space.

\begin{figure*}[ht]
\begin{center}
\small
\newcommand{\height}{2.1cm}
\begin{tabular}{cccccc}
\rotatebox{90}{AvatarStudio} &
    \includegraphics[height=\height]{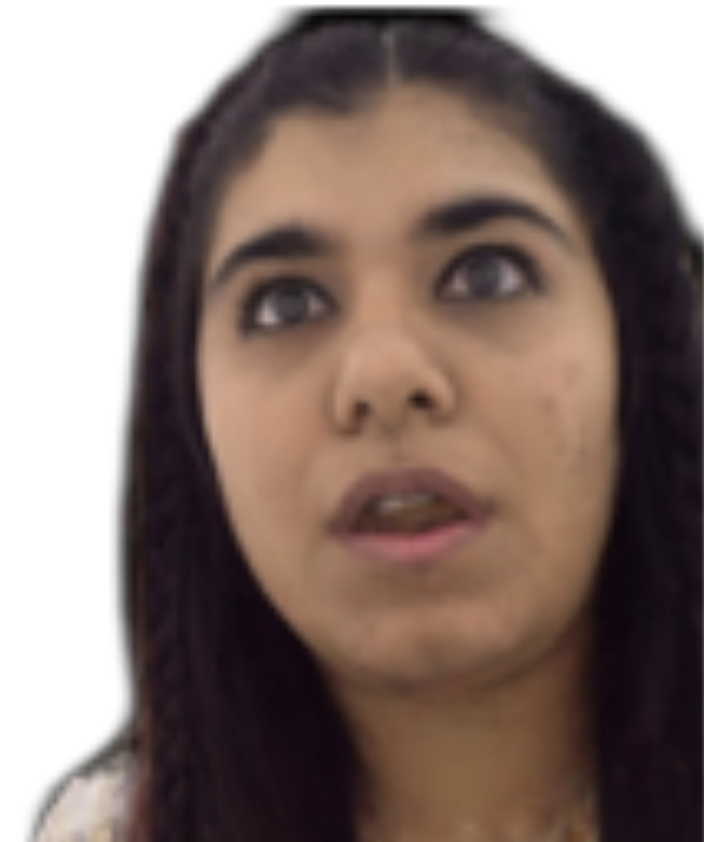} &
    \includegraphics[height=\height]{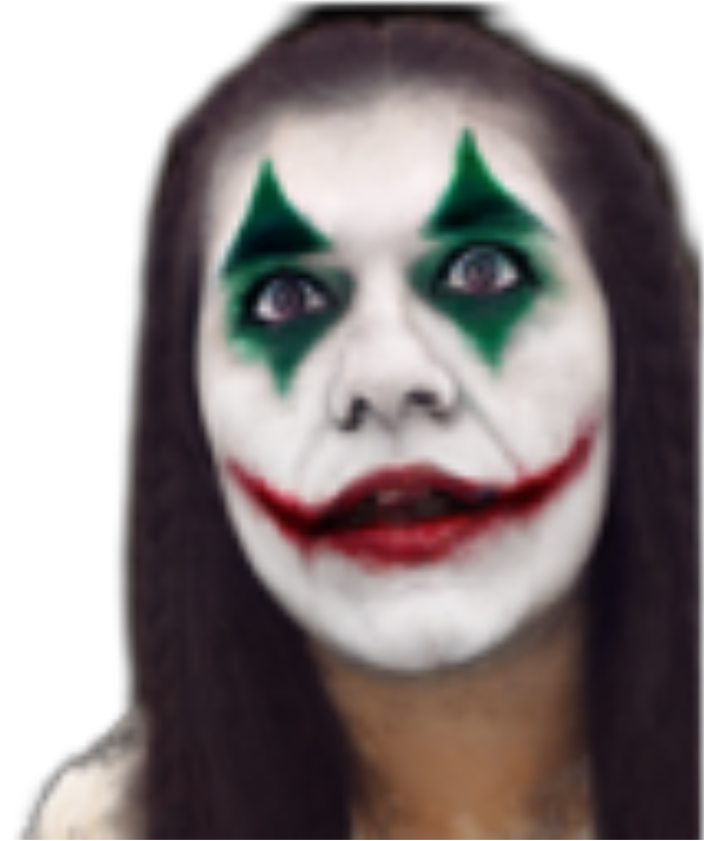} &
    \includegraphics[height=\height]{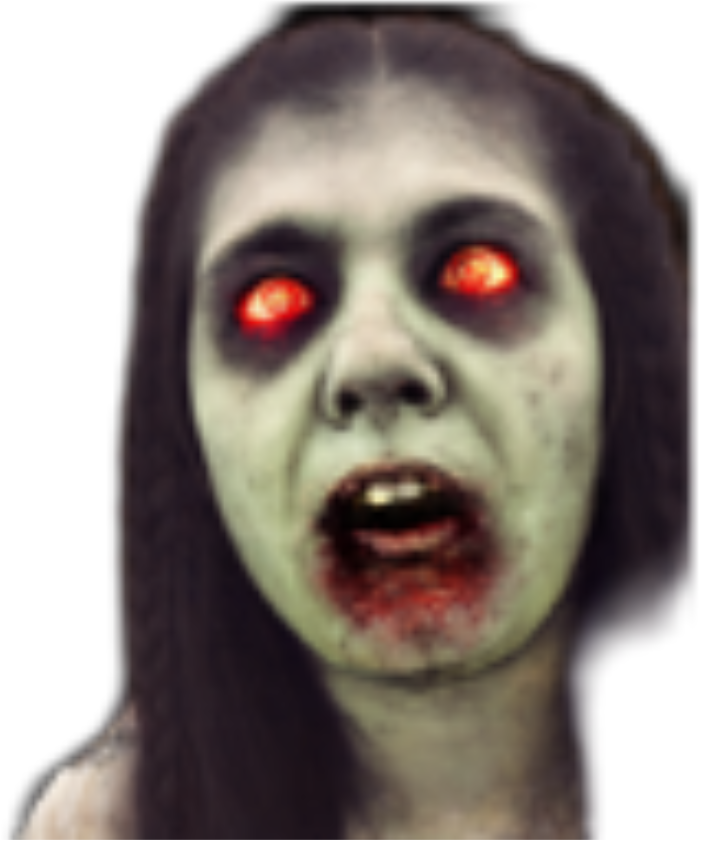} &
    \includegraphics[height=\height]{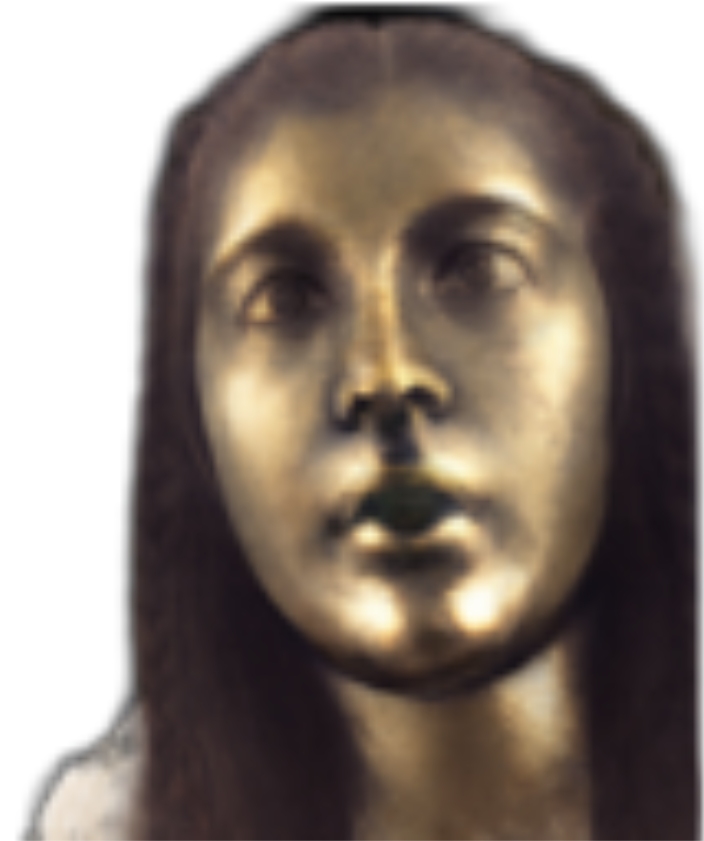} &
    \includegraphics[height=\height]{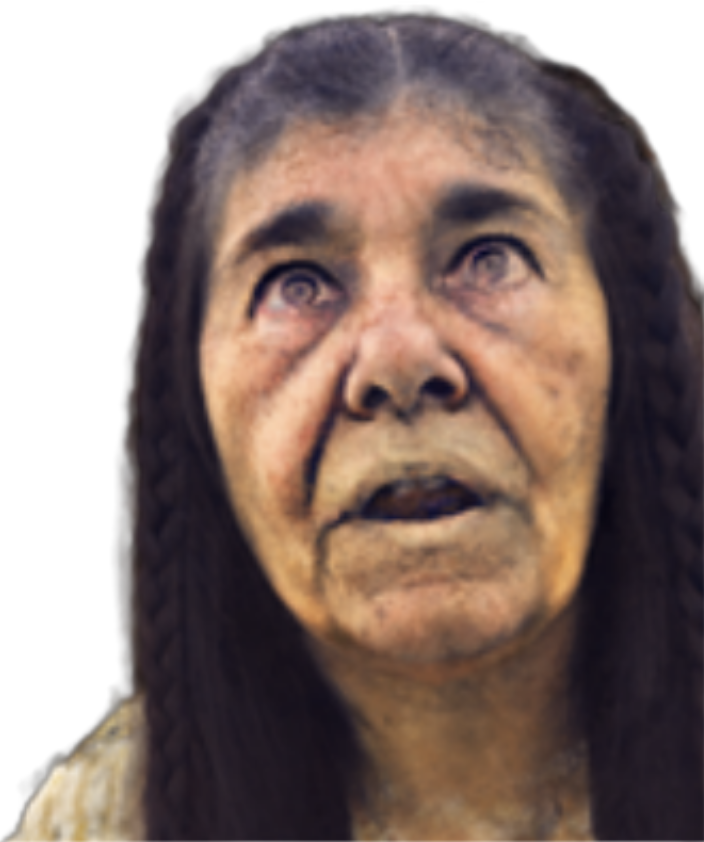} \\
\rotatebox{90}{Ours} &
    \includegraphics[height=\height]{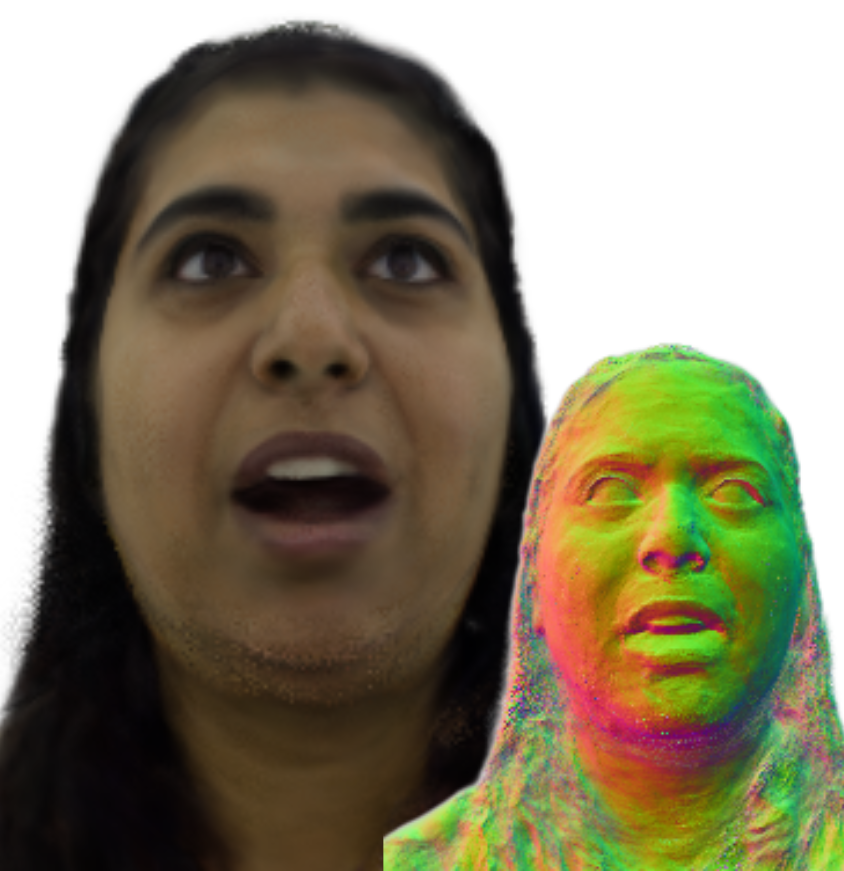} &
    \includegraphics[height=\height]{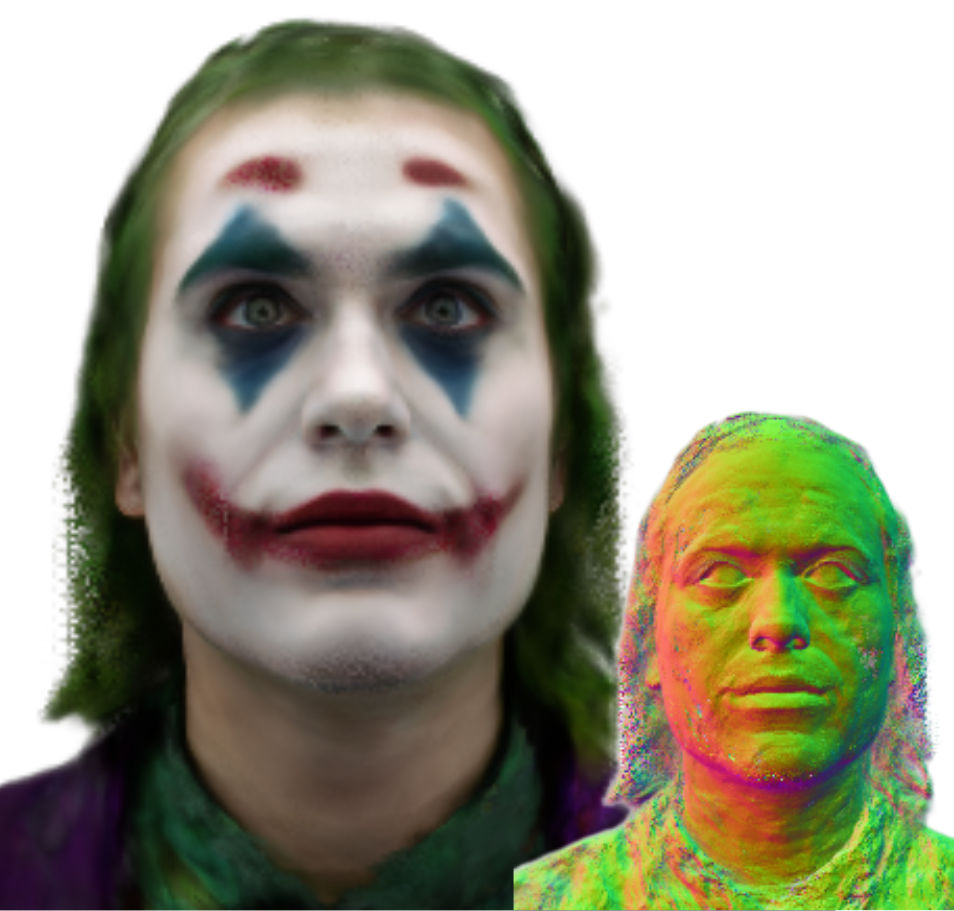} &
    \includegraphics[height=\height]{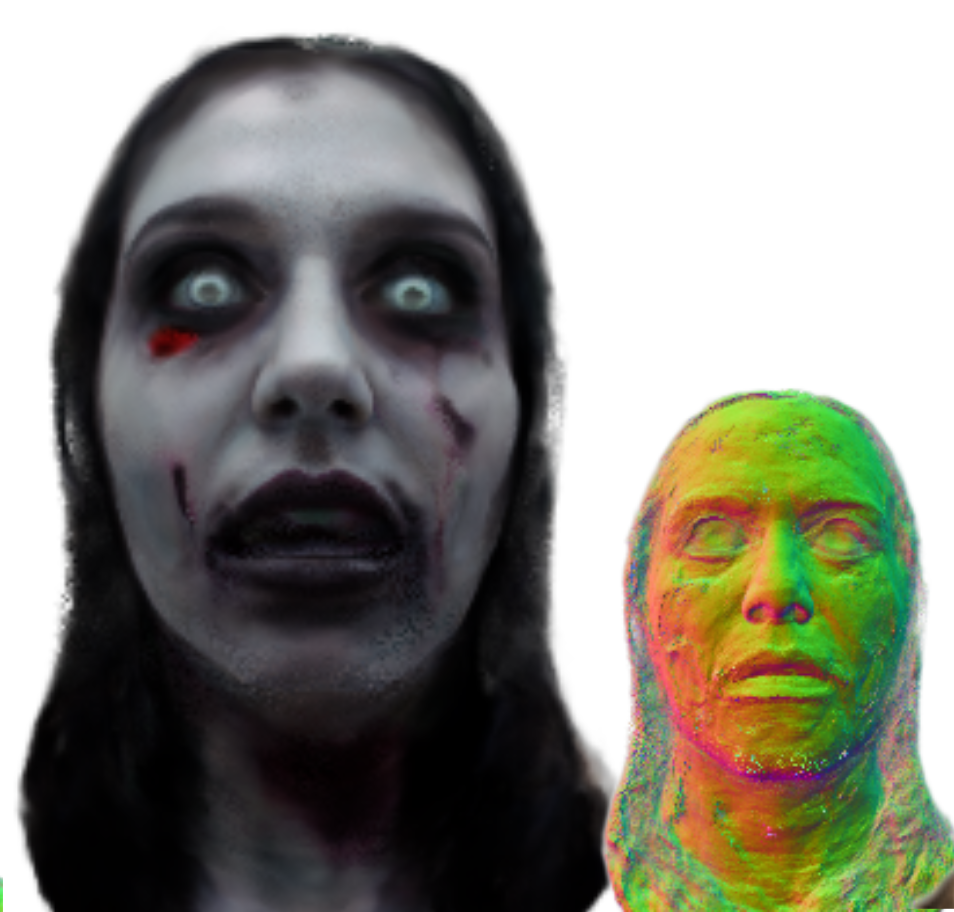} &
    \includegraphics[height=\height]{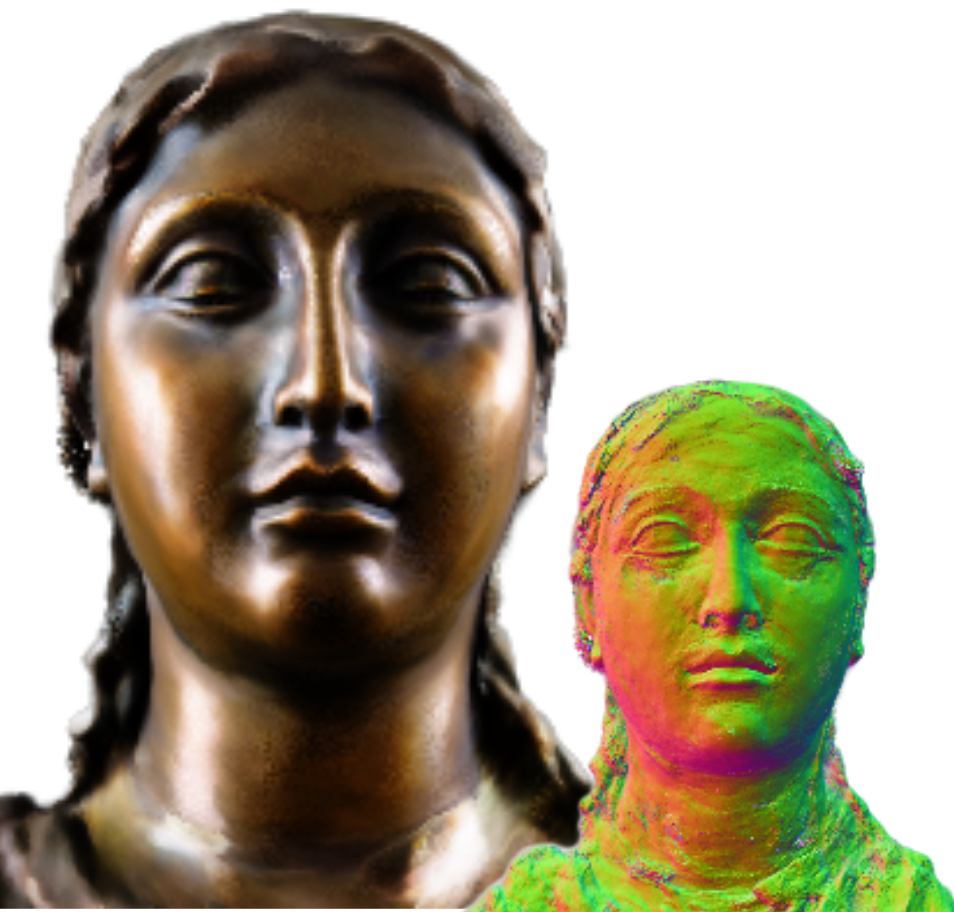} &
    \includegraphics[height=\height]{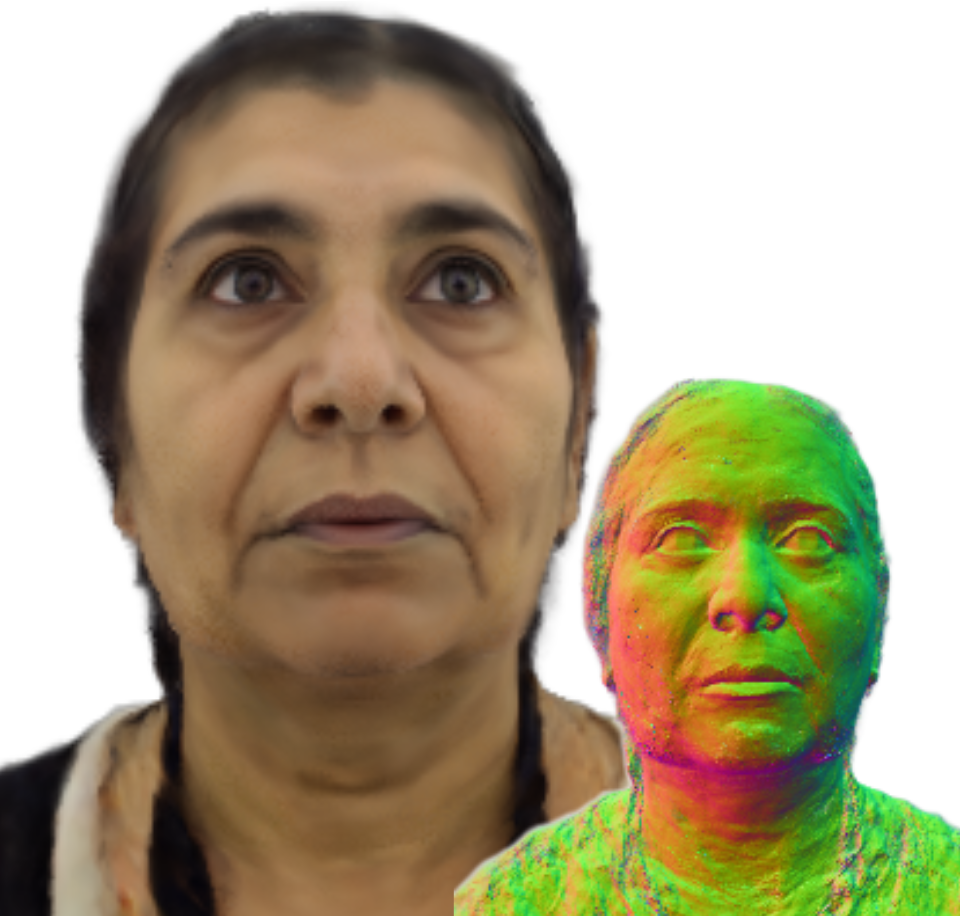} \\
 & Talking & The Joker & Scary Zombie & Bronze Statue & Old Person \\
\end{tabular}
\end{center}
\caption{\label{fig:comparison}We compare our method with AvatarStudio \cite{mendiratta2023avatarstudio}. Please zoom in to see more details.
}
\end{figure*}

\subsubsection{Non-personalized generation with text prompt}\label{sec:famous}

\setlength{\tabcolsep}{0pt}
\begin{figure*}[t!]
\begin{center}
\small
\newcommand{\height}{1.85cm}
\newcommand{\smallheight}{1cm}
\begin{tabular}{ccc | ccc}
\includegraphics[height=\height]{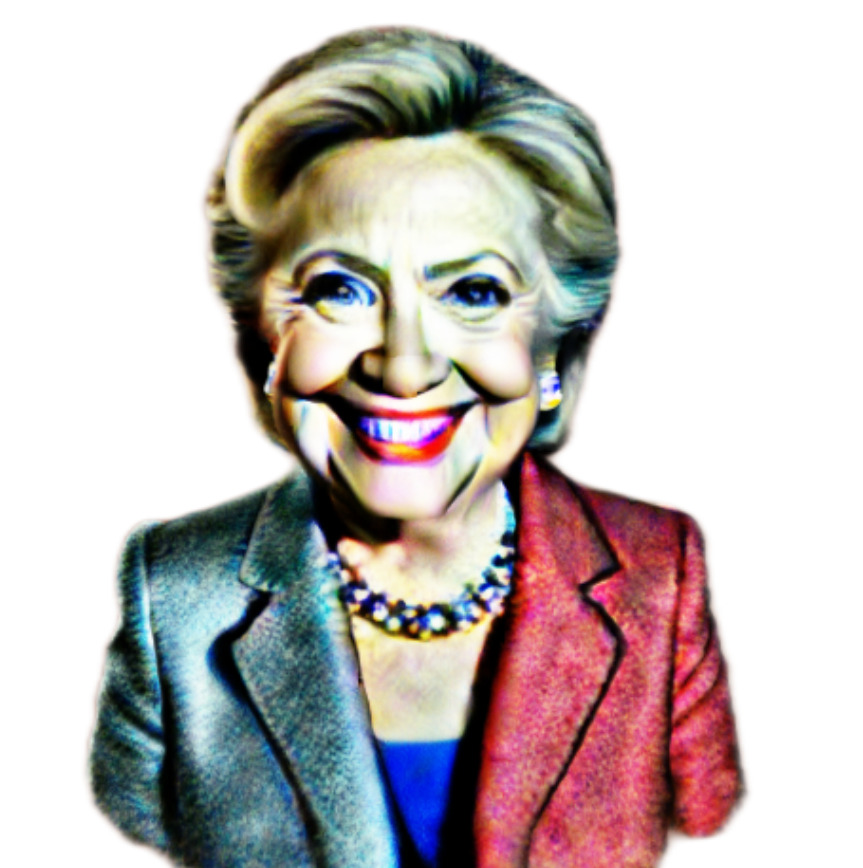} &
    \includegraphics[height=\height]{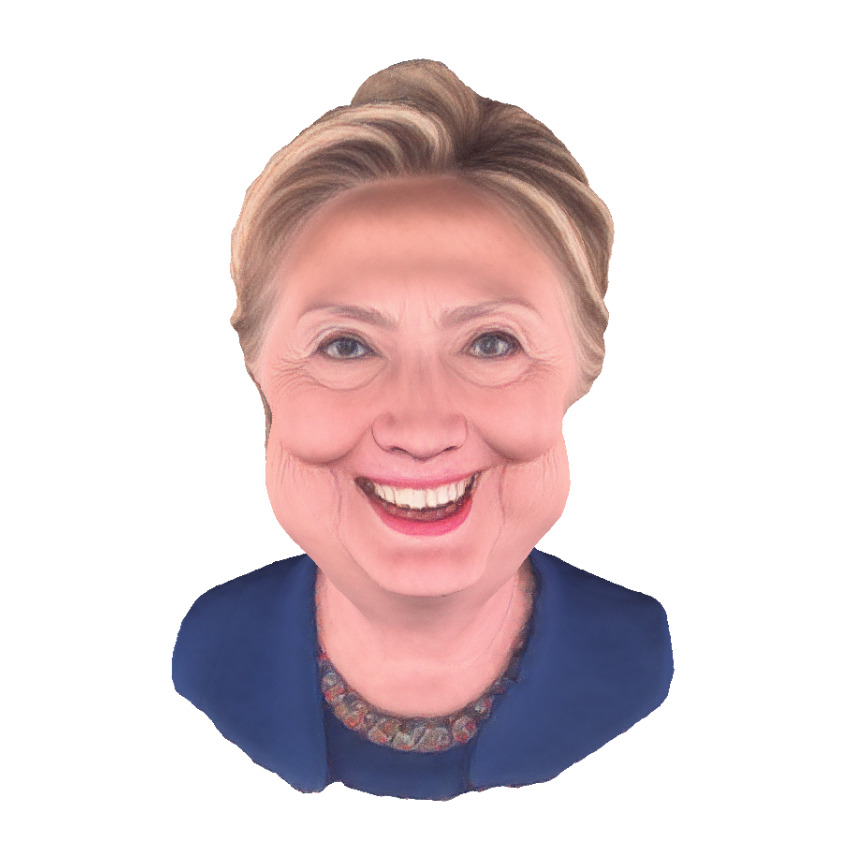} &
    \includegraphics[height=\height]{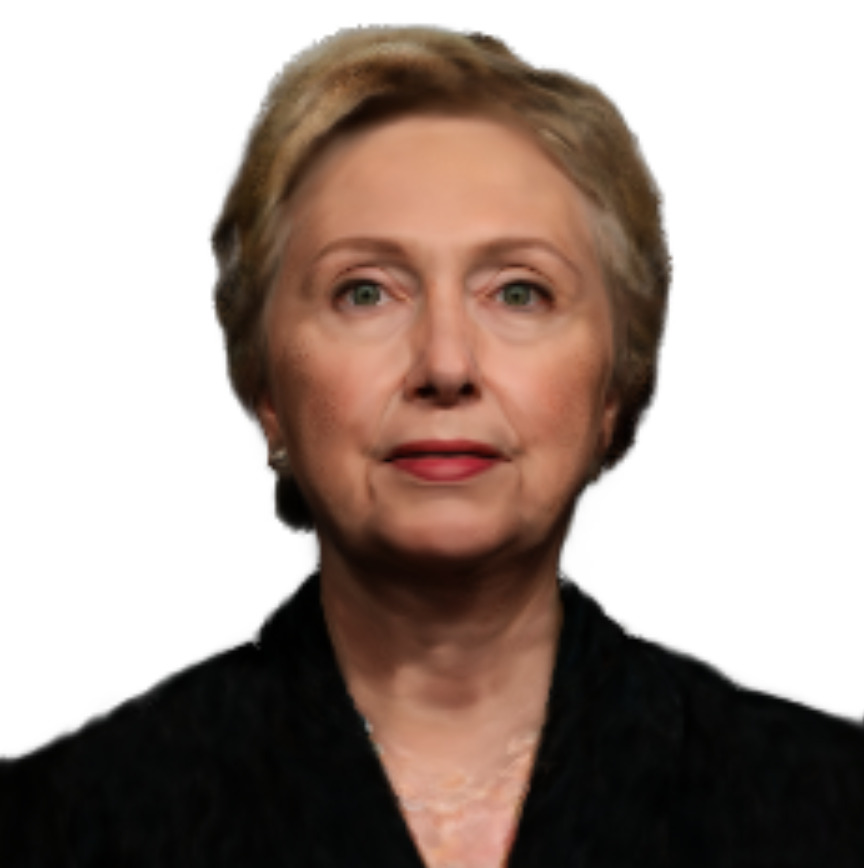} &
    \includegraphics[height=\height]{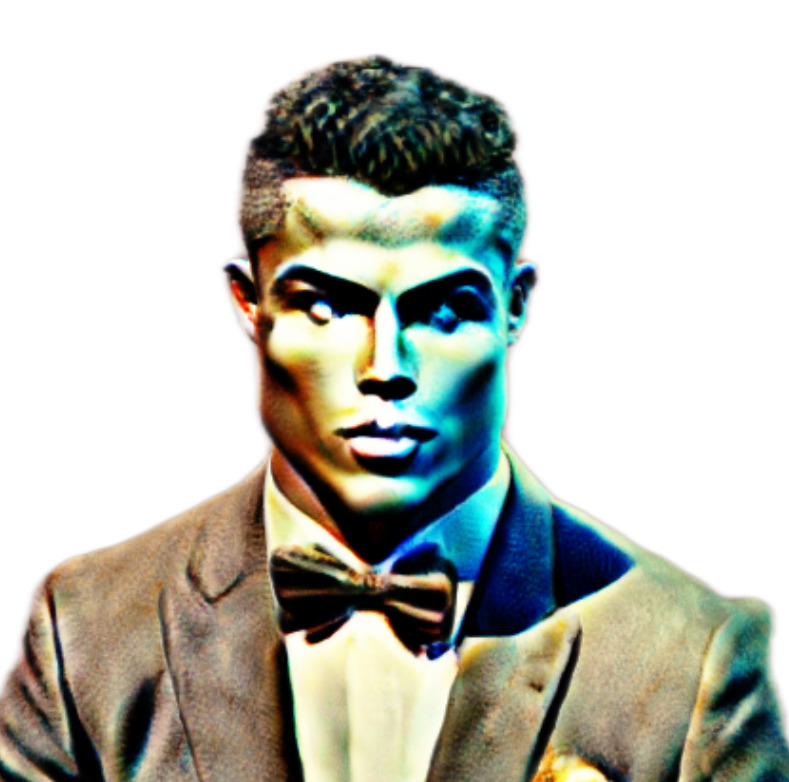} &
    \includegraphics[height=\height]{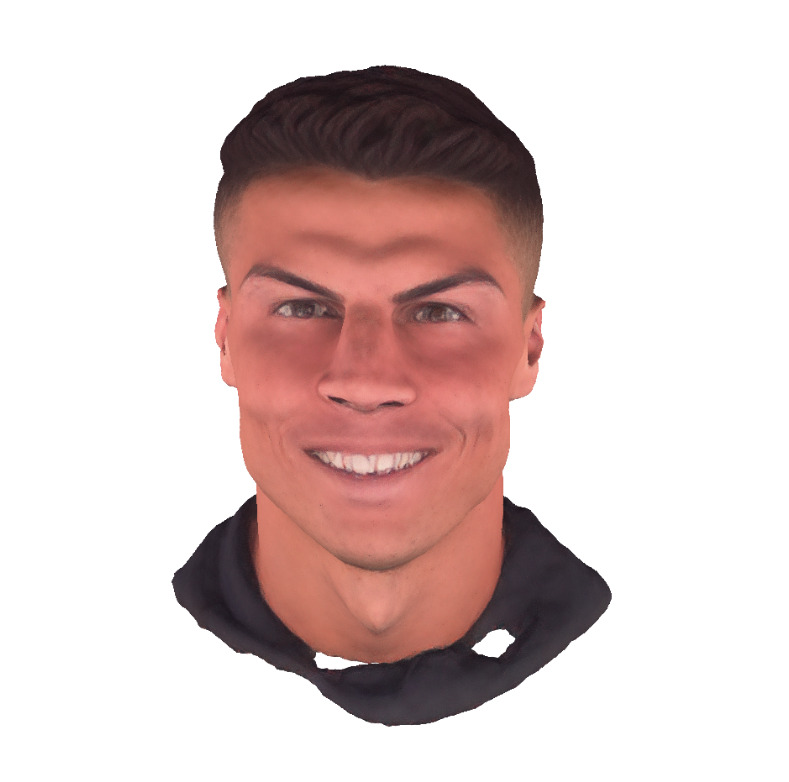} &
    \includegraphics[height=\height]{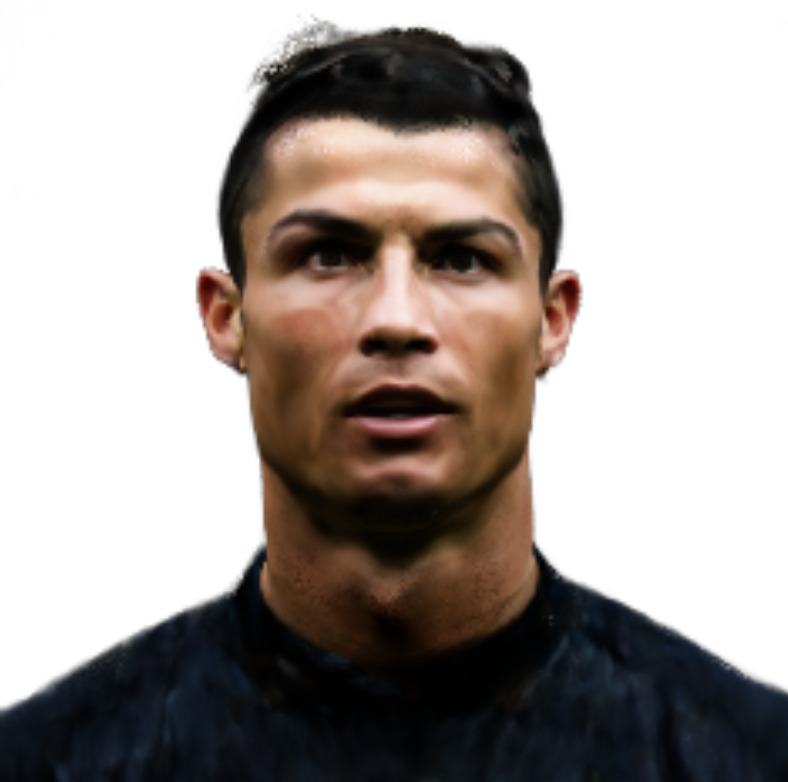} \\ [-3pt]
\includegraphics[height=\smallheight]{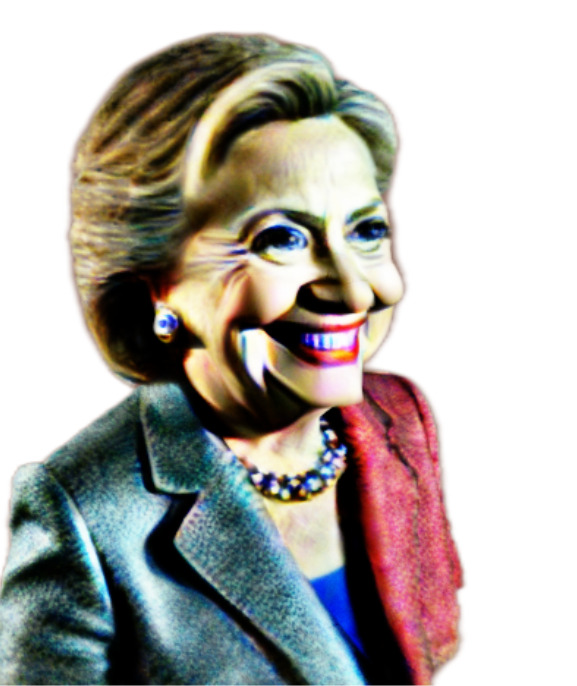}
    \includegraphics[height=\smallheight]{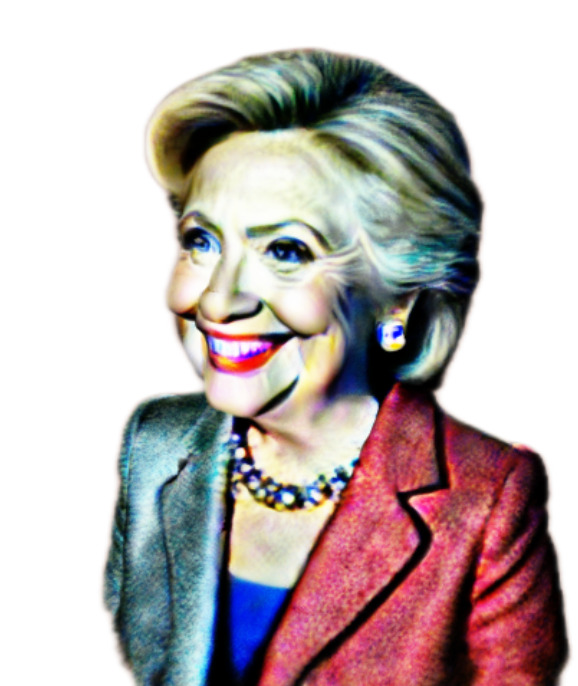} &
    \includegraphics[height=\smallheight]{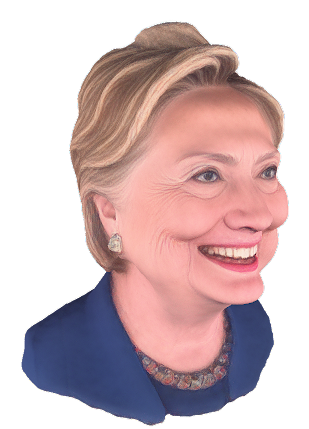}
    \includegraphics[height=\smallheight]{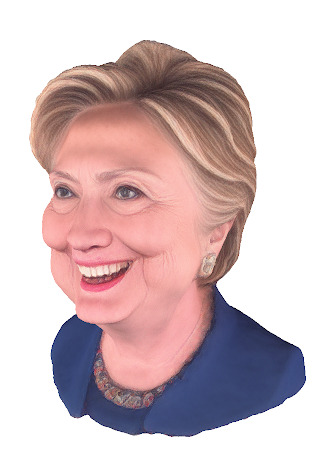} &
    \includegraphics[height=\smallheight]{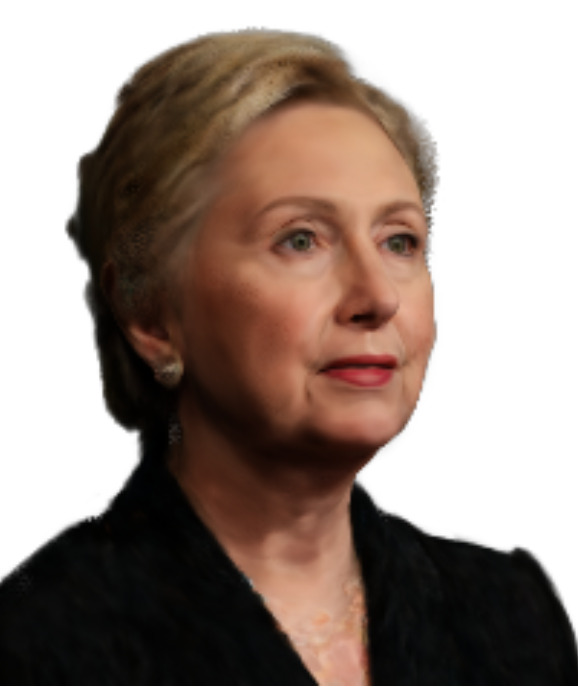}
    \includegraphics[height=\smallheight]{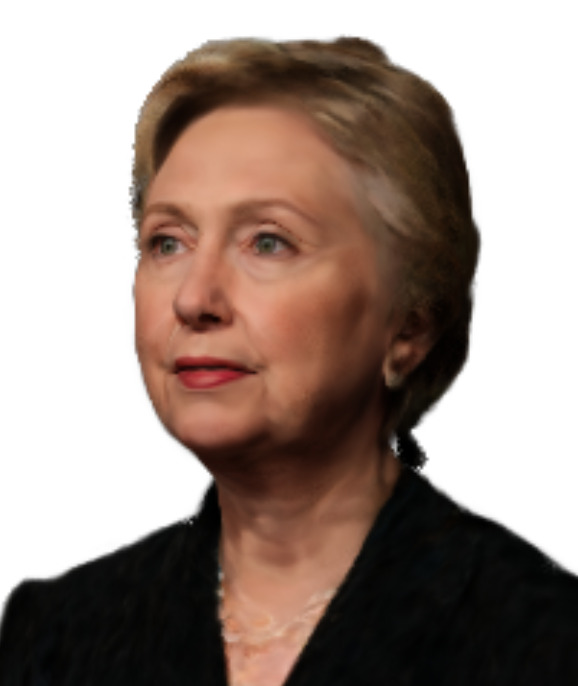} & \hspace{2pt}
\includegraphics[height=\smallheight]{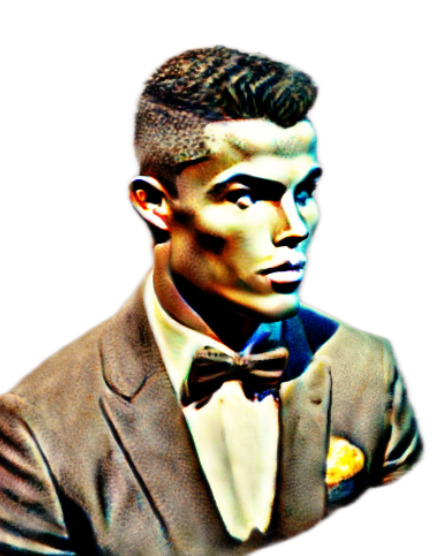}
    \includegraphics[height=\smallheight]{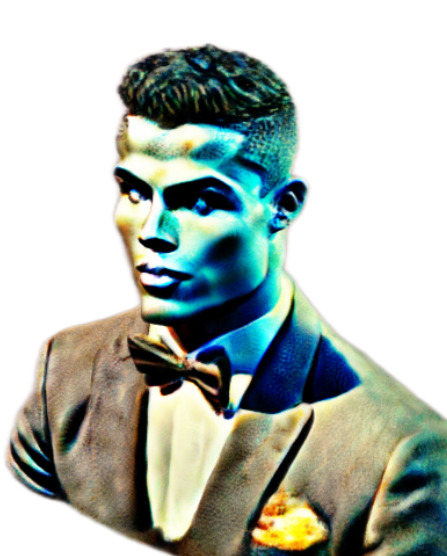} &
    \includegraphics[height=\smallheight]{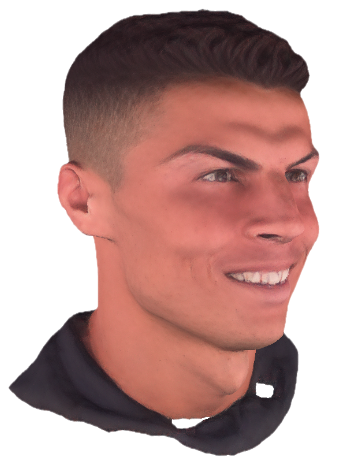}
    \includegraphics[height=\smallheight]{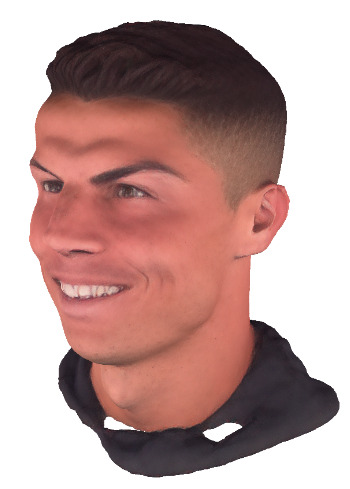} &
    \includegraphics[height=\smallheight]{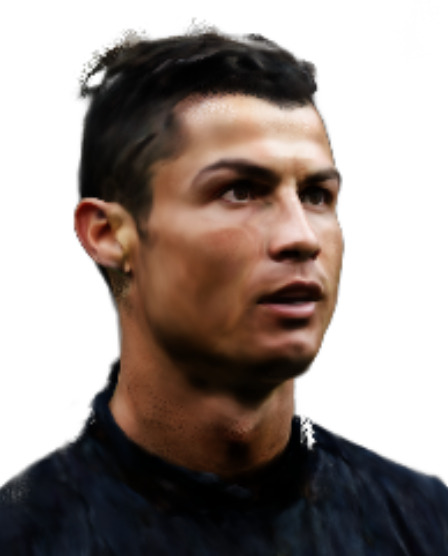}
    \includegraphics[height=\smallheight]{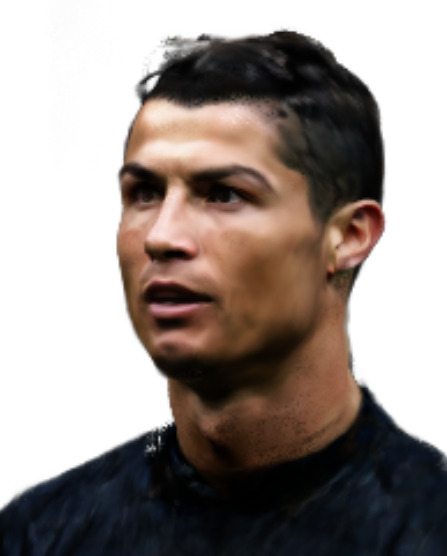} \\ [2pt]
    \hline
\includegraphics[height=\height]{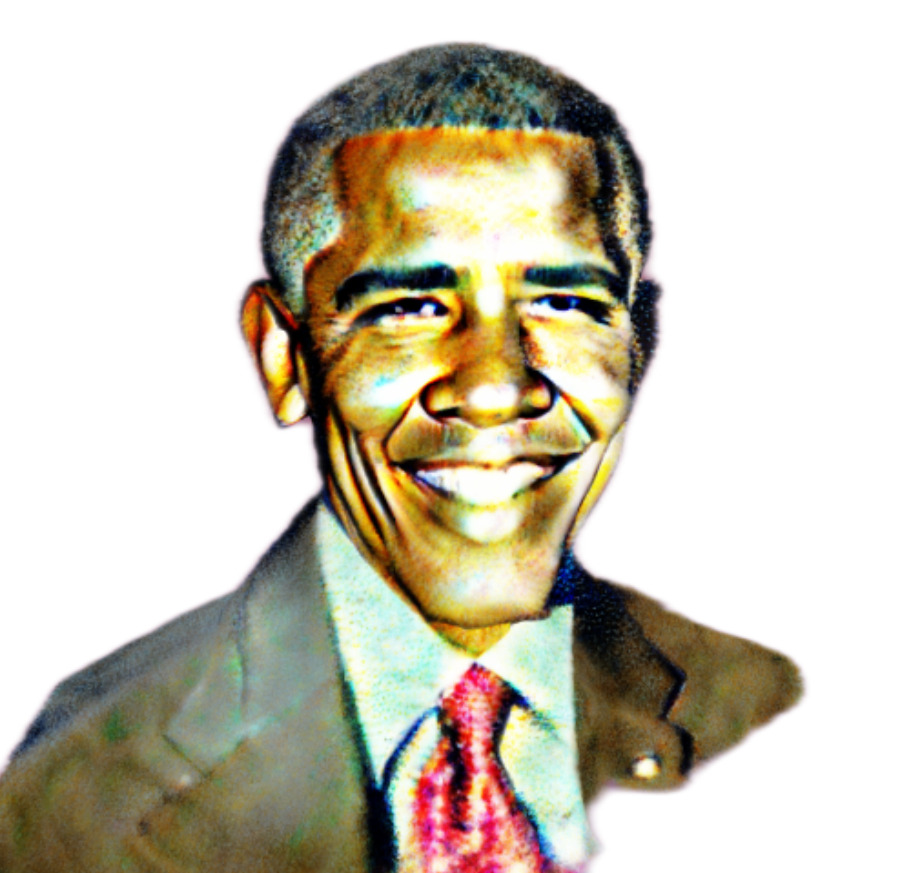} &
    \includegraphics[height=\height]{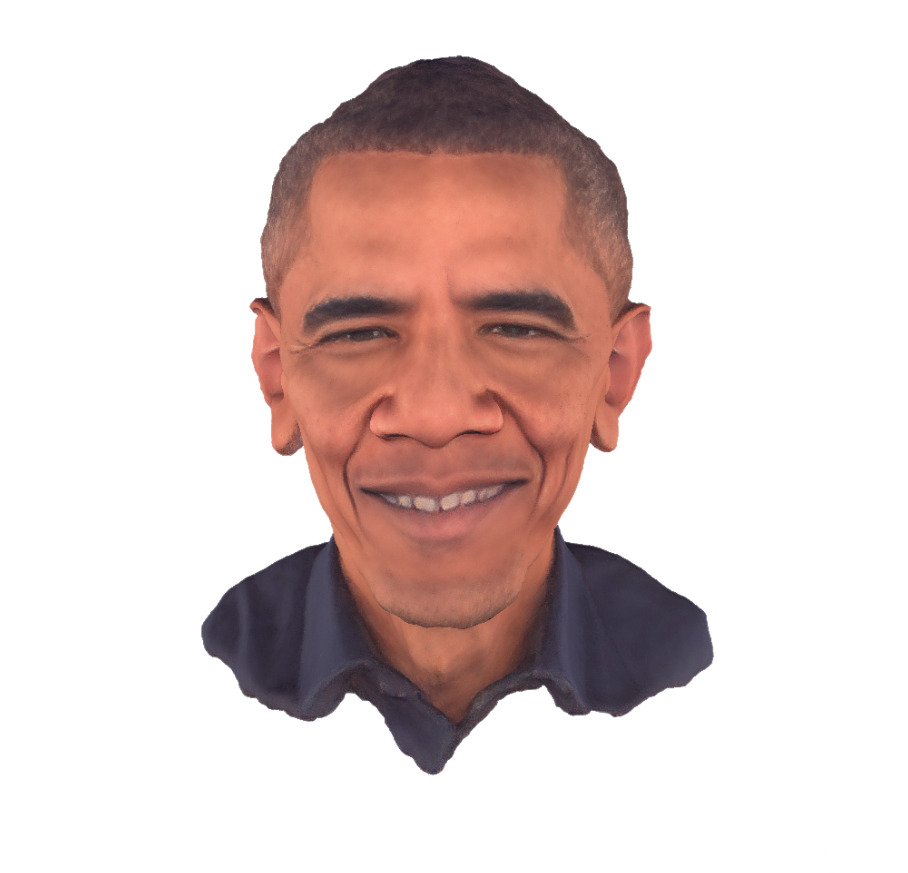} &
    \includegraphics[height=\height]{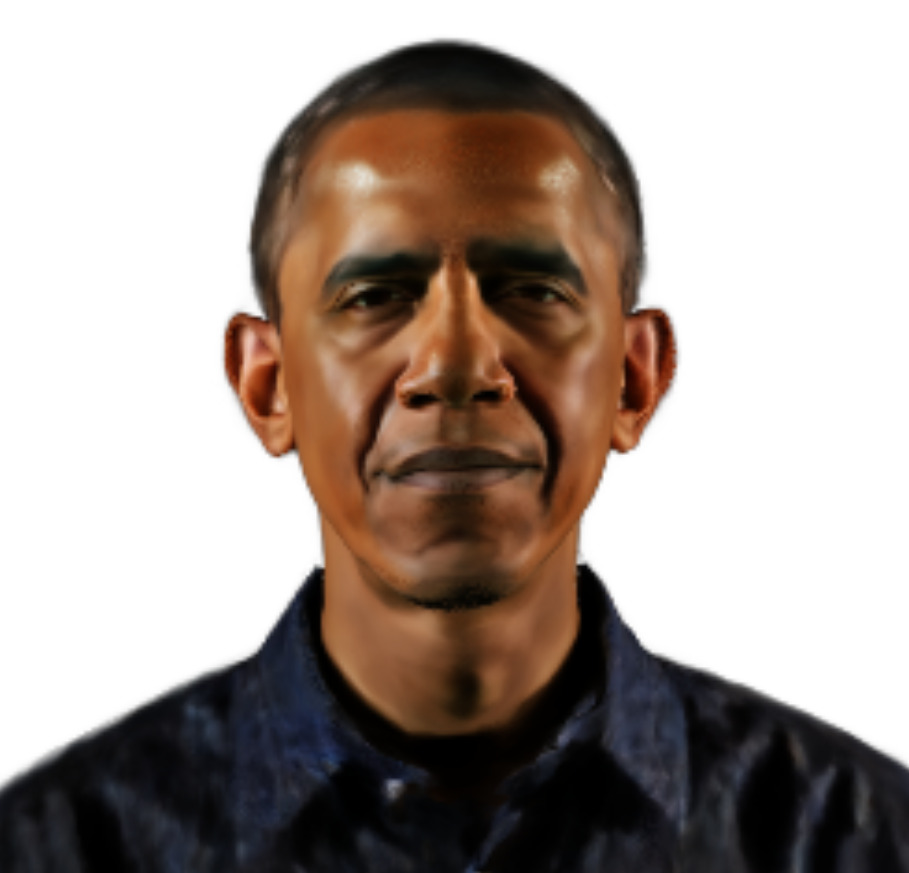} &
    \includegraphics[height=\height]{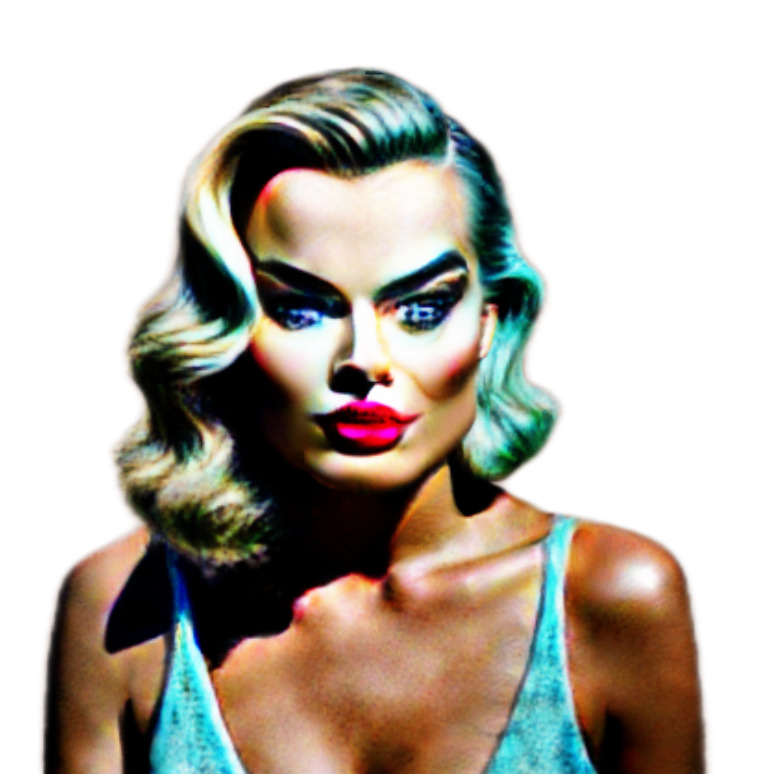} &
    \includegraphics[height=\height]{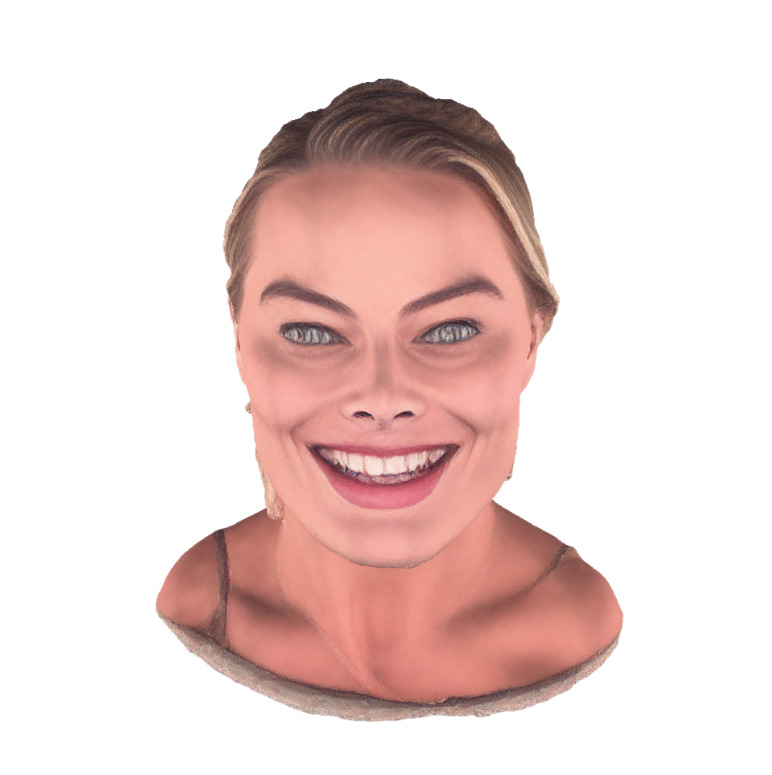} &
    \includegraphics[height=\height]{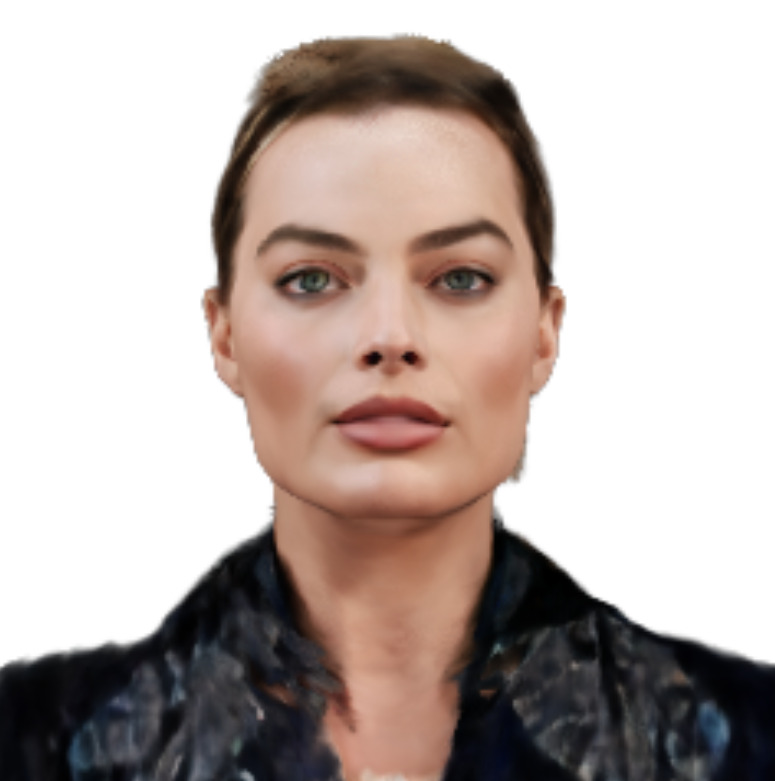} \\
\includegraphics[height=\smallheight]{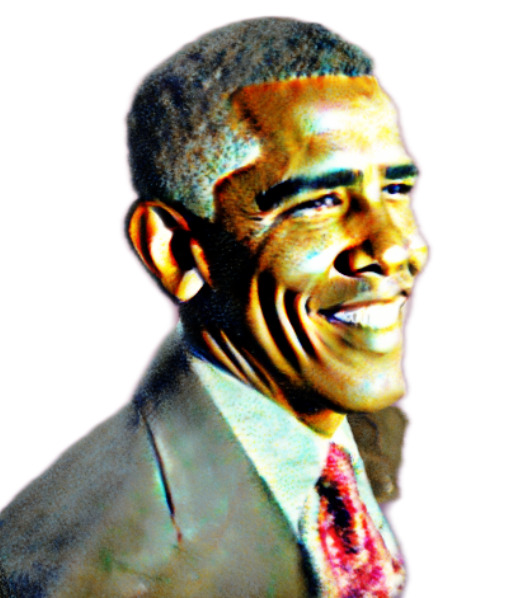} \hspace{-1em}
    \includegraphics[height=\smallheight]{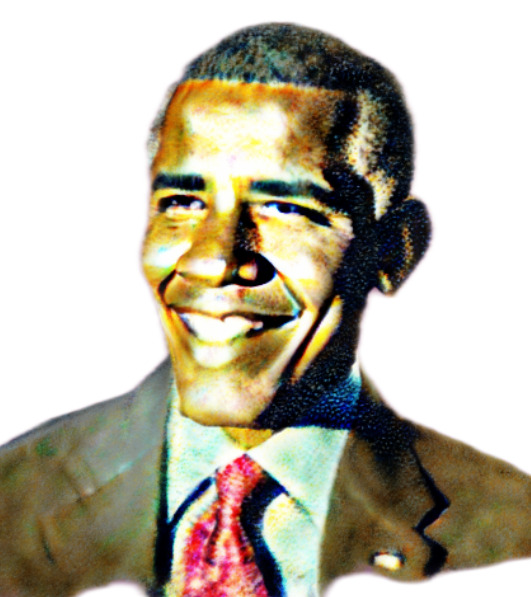} &
    \includegraphics[height=\smallheight]{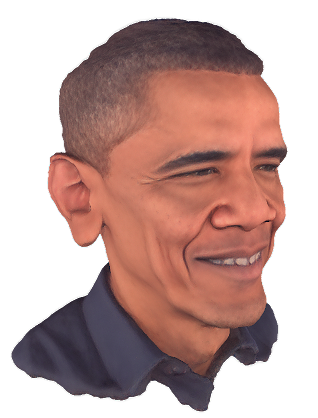} 
    \includegraphics[height=\smallheight]{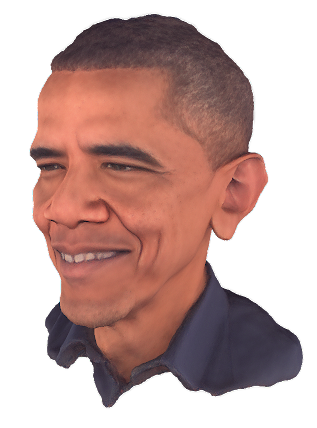} &
    \includegraphics[height=\smallheight]{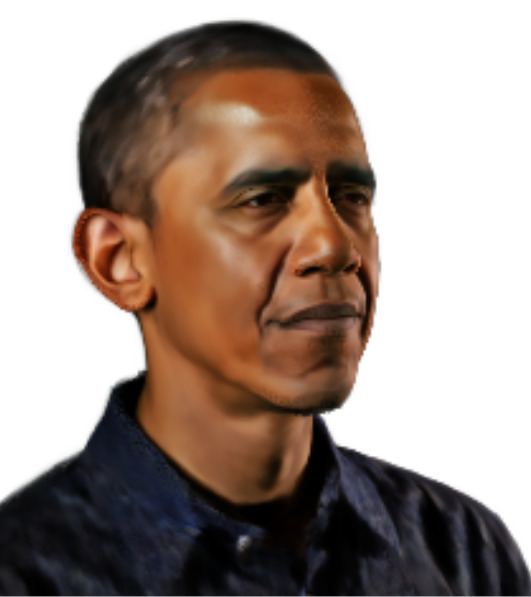}
    \includegraphics[height=\smallheight]{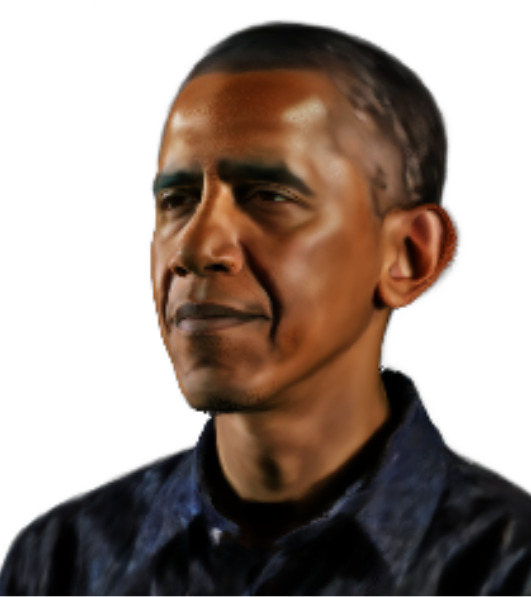}  &
\includegraphics[height=\smallheight]{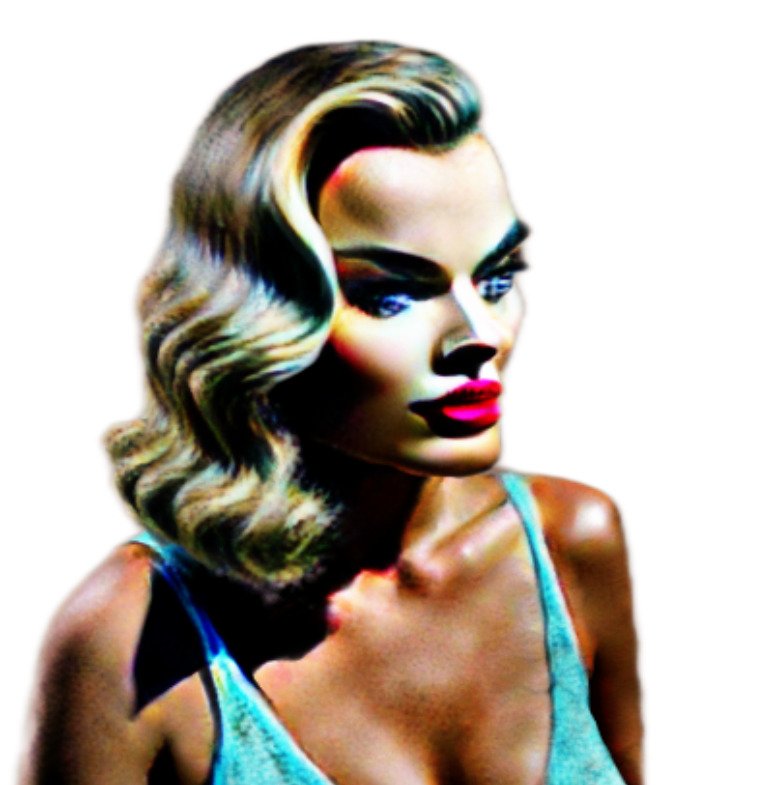} \hspace{-1em}
    \includegraphics[height=\smallheight]{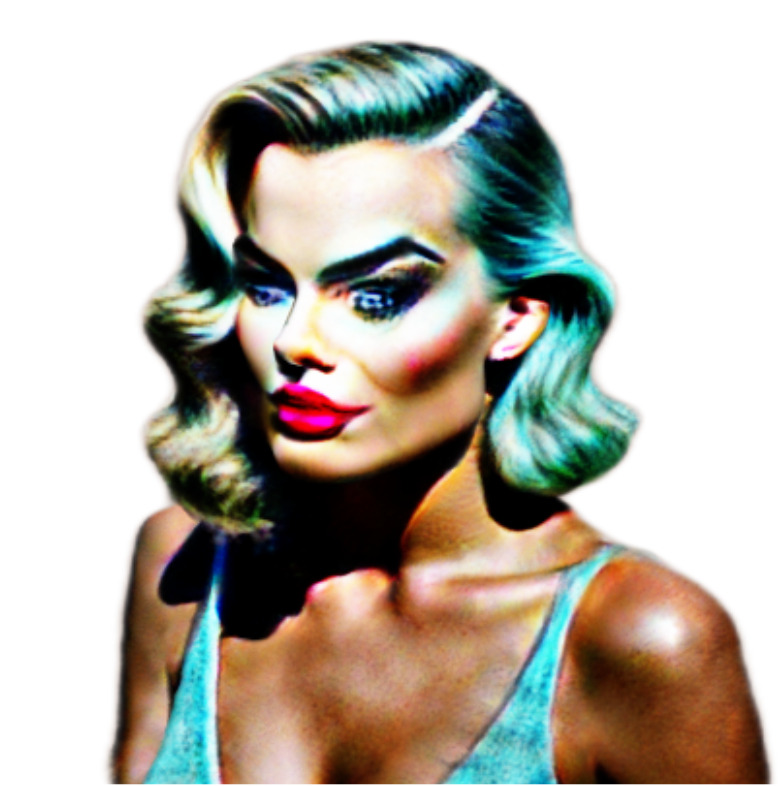} &
    \includegraphics[height=\smallheight]{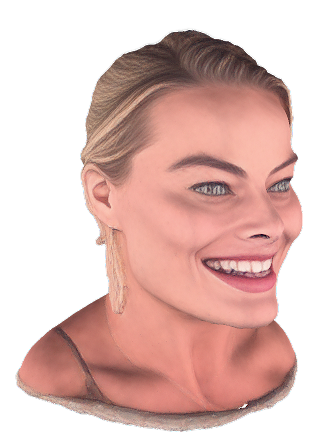}
    \includegraphics[height=\smallheight]{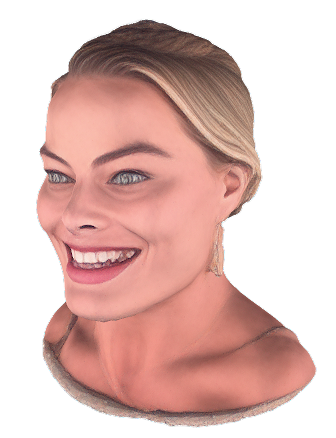} &
    \includegraphics[height=\smallheight]{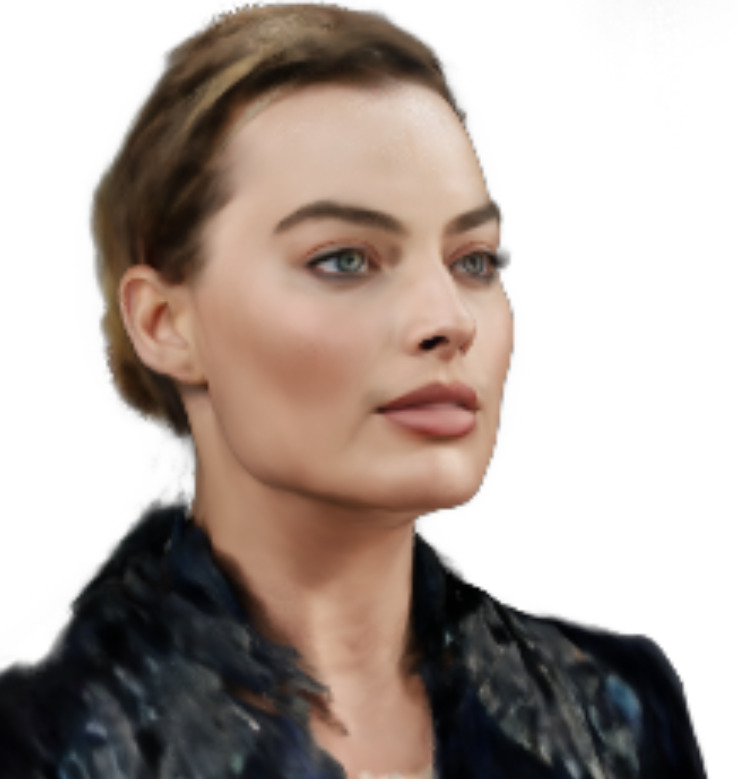}
    \includegraphics[height=\smallheight]{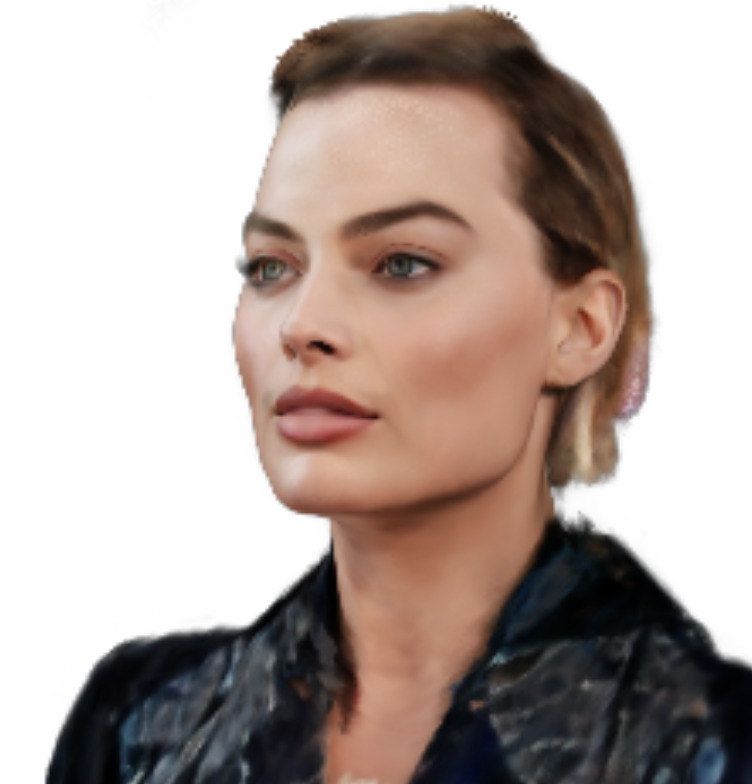} \\ [2pt]
MVDream & HumanNorm & Ours & MVDream & HumanNorm & Ours\\
\end{tabular}
\end{center}
\caption{\label{fig:comparison2} We compare with MVDream \cite{shi2023mvdream} and HumanNorm \cite{huang2023humannorm}. MVDream suffers from color over-saturation and HumanNorm struggles with teeth and eyes and yields a cartoon-ish result. 
}\label{fig:characters}
\end{figure*}

\setlength{\tabcolsep}{6pt}

We now switch to celebrity avatar generation using prompt only, and compare with two recent Text-to-3D methods, MVDream \cite{shi2023mvdream} and HumanNorm \cite{huang2023humannorm}.
MVDream integrates multi-view attention mechanism in Stable Diffusion and fine-tunes the model on large 3D datasets Objaverse. It aims to generate 3D assets of various categories, including humans.
HumanNorm focuses on human avatar generation, which fine-tunes Stable Diffusion on depth and normal renderings of 3D human models to guide the geometry generation.
Both of them employ DMTet \cite{dmtet} to extract mesh geometry to have better decoupling of geometry and texture, and start the test-time optimization from scratch.
The results are visually compared in Fig.~\ref{fig:characters}.
We see that \model achieves much higher realism and fidelity.

\section{Limitations, Impact, and Conclusion}\label{sec:conclusions}
Although we do not require large scale 3D human data, collecting them for hundreds or thousands of subjects can still be a relatively expensive and time consuming effort.
From another point of view, the data we used to constrain the solution space also limit us in the sense that certain extremely out-of-distribution modification is hard to achieve.
Our approach can be also limited by the computational resources, since we need multiple Text-to-Image Diffusion Models, at least each for color and normal, and more if we want to perform mixture of concepts.
Future research can be invested in more modular design and more direct approach to achieve fast and efficient generation and editing.

For its wider adoption, as with all other technologies, we must ensure its development and application to satisfy the security and privacy of the user, and minimize any negative social impact.
In particular, we believe the alignment of the pretrained large Text-to-Image Diffusion Models with human values is becoming ever more important given their growing capability and popularity.

We have presented \model, a next generation text guided 3D avatar generation and editing framework.
Through constraining the solution space, looking for a good geometric prior and choosing a good test-time optimization objective, we have achieved a new level of visual quality, diversity and faithfulness.
The effectiveness of each component has been demonstrated in our thorough ablation and comparison study.
We believe we have made an important step towards an avatar system that people will find it easy and fun to use.

\clearpage

\bibliographystyle{splncs04}
\bibliography{main}

\clearpage
\appendix
\noindent In this supplemental material, we provide further details about our experiments and implementation in Sec.~\ref{sec:app-implementation}. We show additional results in Sec.~\ref{sec:app-more-results}, including both generation and personalized editing results. Finally, in Sec.~\ref{sec:app-failure} we discuss failure cases of our model, along with our hypotheses regarding their causes. As part of the supplemental material, we also provide a web page with more results in video format.

\section{Experimental \& Implementation Details}\label{sec:app-implementation}

\subsection{NeRF Backbone}

Our NeRF is a conditional extension of MipNeRF-360 \cite{mipnerf360} and follows the architecture of Preface \cite{buhler2023preface}. It consists of a proposal (coarse) MLP and a NeRF (fine) MLP. The proposal MLP has 4 layers, and the NeRF MLP has 8 layers. The layer width is $768$ for the proposal and $1536$ for the NeRF MLP. The inputs are encoded using integrated positional encoding with 12 levels for the positional inputs and 4 levels for the view directions.

We initialize the weights of our model after pretraining on 1,450 subjects with a neutral facial expression. Please refer to Preface \cite{buhler2023preface} for a detailed description of the training dataset and training procedure.

To enable faster rendering and lower memory consumption during avatar editing, we only employ a single pass through both the proposal MLP and the NeRF MLP and reduce the number of samples. The proposal MLP is sampled 64 times and the NeRF MLP is sampled 32 times per ray. We find this configuration a good tradeoff between resource usage and rendering quality. If needed, the rendering quality could be improved by employing two proposal passes and quadrupling the number of samples per ray (as in Preface).

\subsection{DreamBooth}

Personalized diffusion prior is a crucial ingredient of our architecture, which employs the geometry prior across all experiments and utilizes the subject-specific prior in every editing scenario.

\begin{figure}[t]
    \centering
            \includegraphics[width=\textwidth]{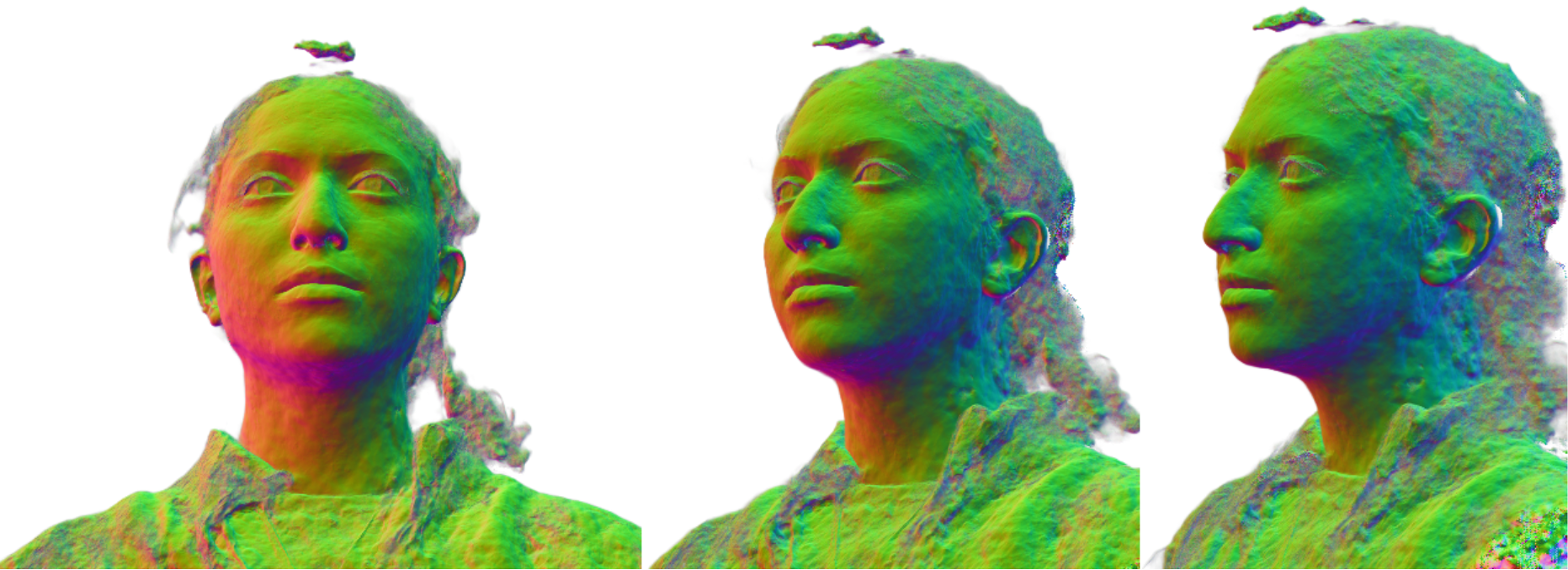}
            \caption{Example of rendered normal maps used to fine-tune the geometry prior.}
            \label{fig:normals_s1}
\end{figure}

For the geometry prior, we utilize a converged avatar model derived from the experimental settings of \cite{buhler2023preface}. Its normal maps are visualized in Fig~\ref{fig:normals_s1}. We render $60$ views from different camera angles and pair them to the same prompt "\textit{A [W] face map of a person}". We fine-tune the model for $800$ iterations at a learning rate of $3e-6$ without regularizing the class prior.

For texture, the personalization process is similar, but it varies based on the type of subject data provided by the user. If the subject's identity is captured via an avatar, we render $60$ RGB views from $3$ different orbits around the head. Conversely, ff the subject's identity is presented through a set of images, we leverage them to fine-tune the diffusion prior. Remarkably, we find that \textbf{our lightweight enrolment process proves to be robust and efficient}, requiring as few as $5$ in-the-wild views to accurately capture the identity. We pair each image with the prompt "\textit{A [V] portrait of a man/woman}" and fine-tune the model for over $400$ iterations with a learning rate of $3e-6$. 

\subsection{Sampling Views for VSD}

In the avatar generation pipeline, selecting appropriate camera angles is crucial to capture the avatar from multiple perspectives. To this end, we design a collection of $1320$ camera orbits around the head, each comprising $30$ uniformly spaced samples. Given that the initialization model is head-shaped, we enrich a subset of these captured views with additional information. During sampling, specific view information is incorporated into the prompt using one of the following descriptors: "\textit{front}", "\textit{side}", "\textit{overhead}", "\textit{low-angle}", "\textit{front}", "\textit{chin}", "\textit{mouth}", "\textit{nose}", "\textit{eyes}", "\textit{hair}" or "\textit{chest}", followed by the term "\textit{view}". We also append "\textit{A DSLR photo of}" at the beginning of each prompt.

It is important to note that this view information is intentionally omitted during the DreamBooth personalization process. The aim is to disregard the influence of specific camera information during personalization.

\subsection{General Hyperparameters}

\begin{figure}[t]
    \centering
            \includegraphics[width=\textwidth]{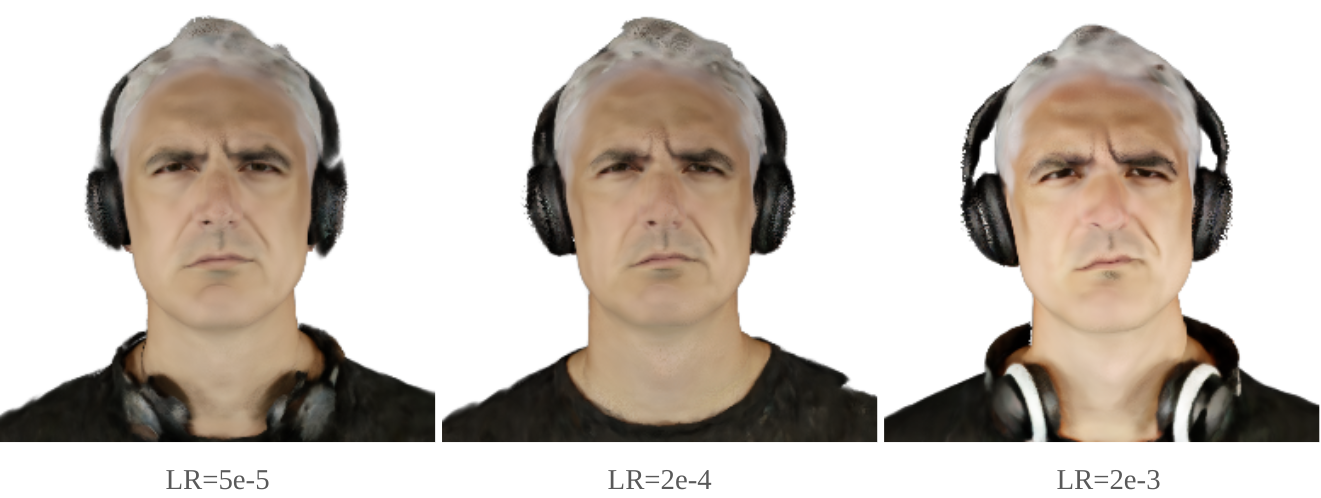}
            \caption{Geometry difference by tuning the NeRF learning rate. Higher learning rates allow the generation of larger volumes.}
            \label{fig:lr}
\end{figure}

The task of avatar synthesis is inherently subjective and visual. The user might prefer different stylization strength, accessory addition, and expression changes. Nonetheless, our approach provides a set of intuitive parameter controls to achieve specific user preferences, as will be detailed further in Sec.~\ref{sec:user guide}.

To approximate the distribution of NeRF-rendered images in Variational Score Distillation (VSD), we update a copy of each diffusion prior at sampling time using a Low Rank Adaptation (LoRA) fine-tuning scheme with a rank of $4$.

In our experiments on non-personalized avatar generation (Sec. 4.4 in the main paper), the major hyper-parameters are set as follows: the Classifier-Free Guidance (CFG) for both texture and geometry priors are set to $3$, with a reduced CFG of $1$ for the NeRF-finetuned model (via LoRA). Given that our target characters are neutral and human-like faces, small adjustments to the initial avatar are sufficient. Thus, we employ a learning rate (LR) of $1e-4$ for all NeRF parameters including latent codes. We also find that for some characters, such as "Will Smith" and "Morgan Freeman", a higher CFG of $5$ can achieve better identity alignment.

Compared to generation, the avatar editing pipeline presents two key differences: Firstly, increasing CFG has a less pronounced effect on the result, which is often preferable. After fine-tuning the diffusion prior to capture a specific identity, the mode-seeking behavior of high CFG sampling adapts better to the underlying distribution of the subject, as its variance is lower than that in the generation case. Therefore, we choose a CFG ranging from $10$ to $25$, keeping the LoRA-adapted prior's CFG at $2$. LR is chosen from a range of $[2e-3, 2e-4]$. The rationale behind these values is explained in Sec.~\ref{sec:user guide}. In all cases, the LoRA-adapted diffusion priors are updated at a LR double that of the avatar's: $\text{LR}_{\text{LoRA}}=2\text{LR}_{\text{NeRF}}$. We find this rule to be crucial to avoid over-saturation and ensure stability around high-quality solutions. Secondly, high-quality results are often obtained earlier in sampling, after around $400$ iterations, though the process may extend to $1000$ iterations to refine details and eliminate artifacts.
However, exceeding this duration, e.g. $2000$ iterations with high LR, can lead to distorted geometry. Therefore, early-stopping or LR decay strategies are essential to the result quality. 

Our method \textbf{requires minimal regularizations} directly applied to the 3D implicit representation. The only regualization is a $\ell^1$ penalization to the accumulated density per ray, in order to avoid unnecessary density generated outside the avatar.

Throughout our experiments, we sample $t$ within $[0.02, 0.8]$, discarding higher values to avoid large shape transformations.

Finally, for the ablation experiment where only latent codes are sampled, a high LR of $3e-2$ is used to expedite the training process.

\subsection{User Guidance}\label{sec:user guide}

\begin{figure}[t!]
    \centering
            \includegraphics[width=\textwidth]{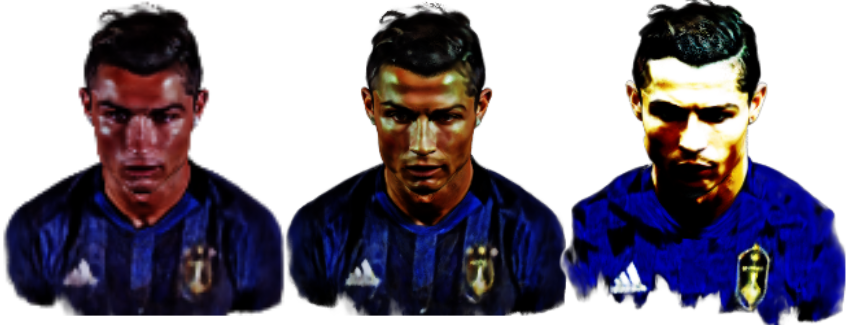}
            \caption{We find that it is crucial to preserve a ratio of $2$ between the LoRA-adapted diffusion priors and the avatar LR. This figure illustrates a set of avatar states along 200 steps of the sampling procedurefor a ratio of $0.3$. While the geometry has converged, the texture is highly saturated and fluctuates.}
            \label{fig:bad-lr-ratio}
\end{figure}

In this section, we list critical factors users must take into account to achieve the desired level of stylization of avatars. 
\begin{itemize}

    \item \textbf{Alignment to text}: It is known that CFG significantly influences diversity and alignment to text prompts. A higher CFG enhances text-alignment at the cost of diversity, while leading to texture loss and over-saturation artifacts similar to Score Distillation Sampling (SDS). According to our experiments, a CFG within the range of $[3, 25]$ produces satisfactory results for non-personalized generation, while editing scenarios favor a range of $[10, 25]$. We are presented with the following tradeoff: Low CFG values yield low-saturation and human-looking textures, while high CFG values yield stronger modifications and more cartoonish results. For non-personalized experiments, where we leverage generic text-to-image diffusion priors, we see a similar tradeoff. However, for generic priors lower CFG values, starting at $3$ also yield high quality results.
    
    \item \textbf{Saturation and stability control}: The most important parameter is the learning rate. In our experiments, we find that it is crucial to keep a $2:1$ ratio between LRs of the LoRA-adapted diffusion priors and the NeRF: $\text{LR}_{\text{LoRA}}=2\text{LR}_{\text{NeRF}}$. A lower ratio yields over-saturated textures and high variability during sampling. In Fig.~\ref{fig:bad-lr-ratio}, we show results of the same avatar with different LR ratios along the sampling procedure. With lower LR ratio, while the geometry is converged, the texture is highly saturated and fluctuated.  

    \item \textbf{Geometry distortion strength}: In our editing experiments, we find it necessary to tune the LR depending on the preferred results. Generally, a high NeRF LR of $\text{LR}_{\text{NeRF}} = 2e-3$ is preferred when larger geometry distortions are expected, such as the addition of a bulky accessory, where $200$ iterations is enough to achieve the desired result and learning rate decay is used for finer detail and smooth geometry. For smaller changes and texture-only stylizations, a lower NeRF LR of $\text{LR}_{\text{NeRF}} = 2e-4$ can be used instead. We can see the effect of the LR in Fig.~\ref{fig:lr}.
    
    \item \textbf{Generic priors for editing}: In the editing pipeline, personalized (identity-aware) priors is used to re-contextualize a particular subject while maintaining the identity. However, if the user may prefer stronger stylizations that even modify some of the identity features (e.g. "\textit{A [V] portrait of the Grinch}"), we find that mixing updates from both personalized priors and generic priors (in a similar fashion as described in Sec. 3.3 (\textit{Mixing and weighting concepts}) in the main paper) can yield more desirable results. Alignment to the identity and the text prompt can be balanced robustly by controlling the update multiplier weights. 
    
    \item \textbf{Texture refinement (optional)}: Though not discussed in our experiments, we find that a few sampling steps with diffusion noise of range $t \in [0.02, 0.5]$ can help refine the texture.
    
\end{itemize}

\subsection{Baselines}
For the sake of brevity, we move the introduction to some of our baselines here. In our quantitative experiments, we also compare against DreamFusion~\cite{poole2023dreamfusion} and Latent-NeRF~\cite{Metzer2022LatentNeRFFS}. As established and highly influential methods for generic 3D generation, they perform inevitably worse than the other two baselines, MVDream~\cite{shi2023mvdream} and HumanNorm~\cite{huang2023humannorm}. However, the comparisons still provide a good perspective of the improvements made by the latest research regarding domain-specific generation of human avatars.

\section{Additional Results}\label{sec:app-more-results}

\begin{figure}[t]
    \centering
        \begin{subfigure}[t]{0.22\linewidth}
            \includegraphics[width=\textwidth]{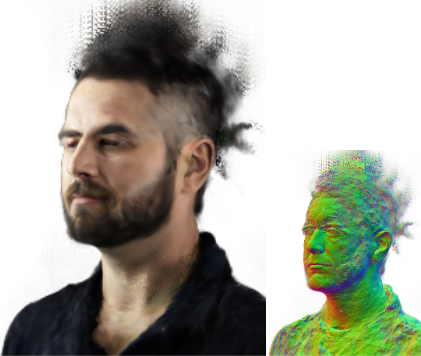}
            \caption{\textit{A DSLR photo portrait of a man with crazy spiky hair}. }
            \label{fig:hair}
        \end{subfigure}
        \begin{subfigure}[t]{0.4\linewidth}
            \includegraphics[width=\textwidth]{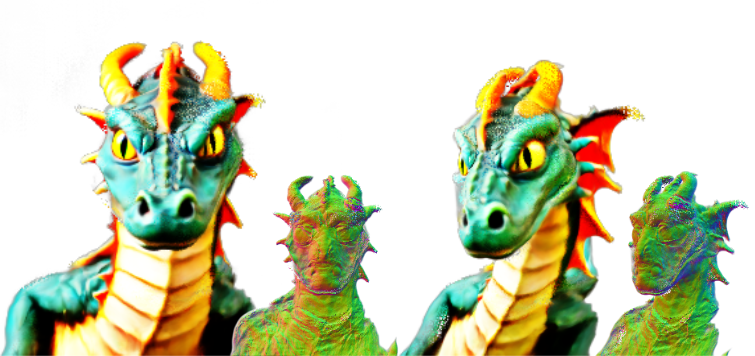}
            \caption{\textit{A DSLR photo of a dragon}. }
            \label{fig:dragon}
        \end{subfigure}
        \begin{subfigure}[t]{0.26\linewidth}
            \includegraphics[width=\textwidth]{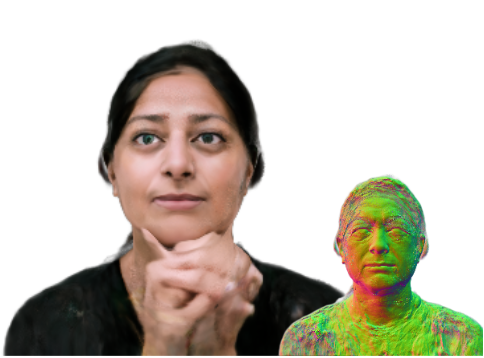}
            \caption{\textit{A DSLR photo portrait of a woman who is showing her hands} }
            \label{fig:hands}
        \end{subfigure}
            \caption{Model shows different failure modes. Here, we highlight the cases of (a) Failure case with undefined shapes such as hair, especially in regions outside of the facial area. (b) Failure case of geometry generation, where traces of humanoid features can be seen in the normal maps despite generating a non-humanoid subject, and (c) Our model fails to generate the volume corresponding to the hands and instead inpaints them.} \label{fig:failure}
\end{figure}

In what follows, we provide more examples of non-personalized generation and editing results to enhance the reader's understanding of our model's performance. Please refer to the supplementary page for more results in video format.

\begin{figure*}[ht]
\begin{center}
\includegraphics[width=\textwidth]{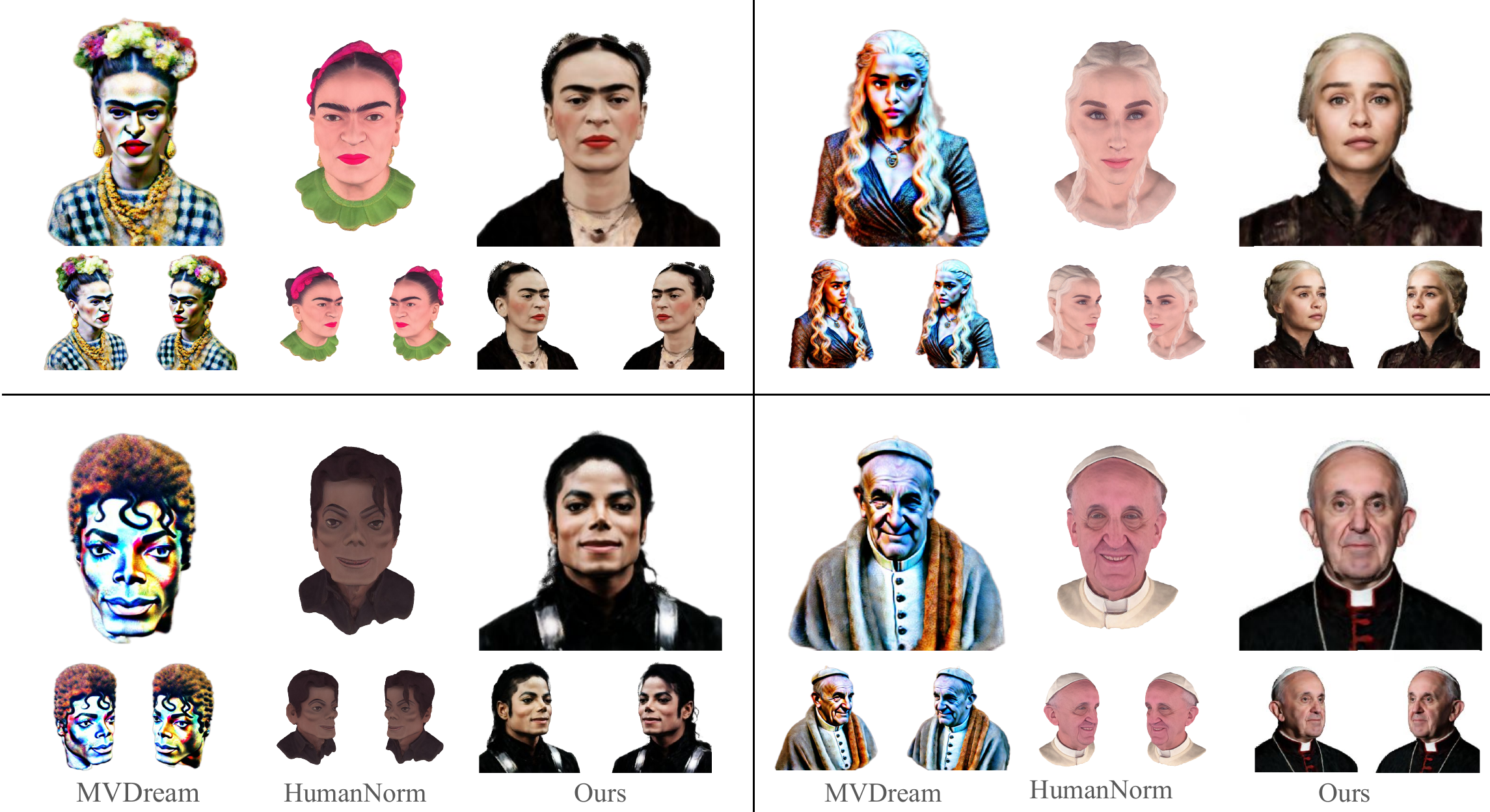}
\end{center}
\caption{\label{fig:supp-comparison2} Comparisons with MVDream and HumanNorm. MVDream suffers from over-saturation and HumanNorm struggles with teeth and eyes and yields cartoon-ish results. }
\end{figure*}

\subsection{More Avatar Generation Results}
In Fig.~\ref{fig:supp-comparison2} we demonstrate our results compared to the main baselines on famous characters (left-right, up-down): \textit{Frida Kahlo}, \textit{Daenerys Targaryen}, \textit{Michael Jackson}, and \textit{Pope Francis}. Once again we see how \model outperforms the baselines in terms of both visual quality and alignment to real identities.

\subsection{More Avatar Editing Results}

\begin{figure*}[ht!]
\begin{center}
\includegraphics[width=\textwidth]{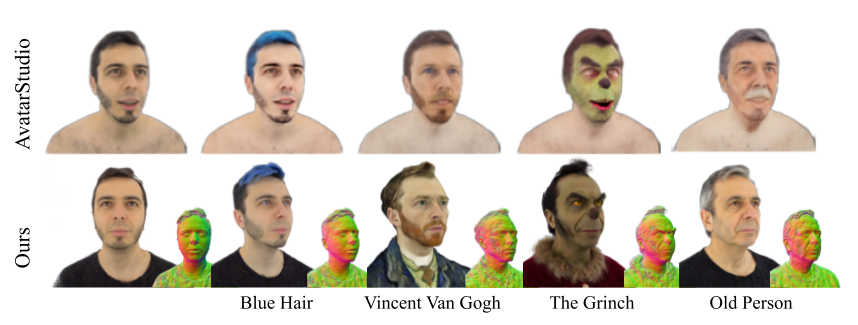} 
\end{center}
\caption{\label{fig:supp-editing}Our novel framework, \model, can successfully change facial expressions, features, and add accessories or specific styles to the person. The top row is directly extracted from a figure AvatarStudio paper. Results show qualitatively how our method yields results that have higher quality and better text alignment than our baseline.
}
\end{figure*}

In Fig.~\ref{fig:supp-editing}, we show another comparison against AvatarStudio~\cite{mendiratta2023avatarstudio}, a method specialized in editing and with no capability for generation. AvatarStudio modifies an existing avatar that already represents the target identity, while our method generates such identity from scratch and then re-contextualizes it by a text prompt. We can see that our method captures the identity correctly while generating drastic changes in geometry and texture, with more realistic results with sharper features. 

\section{Failure Cases}\label{sec:app-failure}

In this section, we discuss several noteworthy instances of failure cases encountered in \model, as depicted in Fig~\ref{fig:failure}.

Firstly, as shown in Fig.~\ref{fig:hair}, generating scenarios like "\textit{A DSLR photo portrait of a man with crazy spiky hair}" presents a dual challenge. Initially, while it may appear as a surface in certain instances, defining hair as a volume proves intricate due to its composition of fine elements. Many methods impose smoothness regularizations on their 3D representations, often leading to the modeling of hair as a textureless surface. Notably, in our approach, we do not impose regularizations to our 3D representation except a density penalization. Consequently, the inherent uncertainty of hair surfaces results in a noisy volume, rendering it unrealistic. Furthermore, the lack of definition is exacerbated by the views sampled during generation. Typically, we focus on sampling views of the facial region to enhance the definition of facial traits and texture, diminishing attention to other areas surrounding the head and compromising generation quality in these regions. Addressing these challenges will be a focus of future research efforts.

Secondly, Fig.~\ref{fig:dragon} illustrates the result corresponding to: "\textit{A DSLR photo of a dragon}". This result offers two insights. Firstly, it demonstrates our model's capability of generating non-humanoid portraits with surprisingly high level of detail and text alignment. However, upon closer examination of the surface normals, we can still see remaining humanoid features. The generative process, initialized with a human appearance, struggles to fully transform all facial features. Nevertheless, this experiment underscores the potential for more generic 3D asset generation using our method.

Lastly, Fig.~\ref{fig:hands} showcases the result corresponding to "\textit{A DSLR photo portrait of a woman showing her hands.} Creating new volumes from scratch with \model is not always successful, particularly for volumes detached from the head. This limitation can be attributed in part to our initialization, which has been trained to adapt to upper-body portrait shapes. When attempting to generate a pair of hands from scratch, our method converges to a local minimum that inpaint hands onto the subject's neck, failing to produce their true volume.

\end{document}